\documentclass[10pt,journal,compsoc]{IEEEtran}
\ifCLASSOPTIONcompsoc
  \usepackage[nocompress]{cite}
\else
  \usepackage{cite}
\fi

\usepackage{epsfig}
\usepackage{graphicx}
\usepackage{amsmath}
\usepackage{amssymb}
\usepackage{adjustbox}
\usepackage{multirow}
\usepackage{color}
\usepackage{fancyhdr}
\ifCLASSINFOpdf
\else
\fi
\begin{document}
\title{Rolling-Unrolling LSTMs for \\Action Anticipation from First-Person Video}

\author{Antonino~Furnari\thanks{Antonino Furnari and Giovanni Maria Farinella are co-first authors.} and
        Giovanni Maria Farinella%
\IEEEcompsocitemizethanks{\IEEEcompsocthanksitem A. Furnari and G. M. Farinella are with the Department of Mathematics and Computer Science, University of Catania, Italy\protect\\
E-mail: \{furnari,gfarinella\}@dmi.unict.it}%
}
\IEEEtitleabstractindextext{%
\begin{abstract}
In this paper, we tackle the problem of egocentric action anticipation, i.e., predicting what actions the camera wearer will perform in the near future and which objects they will interact with. Specifically, we contribute Rolling-Unrolling LSTM, a learning architecture to anticipate actions from egocentric videos. The method is based on three components: 1) an architecture comprised of two LSTMs to model the sub-tasks of summarizing the past and inferring the future, 2) a Sequence Completion Pre-Training technique which encourages the LSTMs to focus on the different sub-tasks, and 3) a Modality ATTention (MATT) mechanism to  efficiently fuse multi-modal predictions performed by processing RGB frames, optical flow fields and object-based features. The proposed approach is validated on EPIC-Kitchens, EGTEA Gaze+ and ActivityNet. The experiments show that the proposed architecture is state-of-the-art in the domain of egocentric videos, achieving top performances in the 2019 EPIC-Kitchens egocentric action anticipation challenge. The approach also achieves competitive performance on ActivityNet with respect to methods not based on unsupervised pre-training and generalizes to the tasks of early action recognition and action recognition. To encourage research on this challenging topic, we made our code, trained models, and pre-extracted features available at our web page: \textit{http://iplab.dmi.unict.it/rulstm}.
\end{abstract}

}

\maketitle

\IEEEdisplaynontitleabstractindextext
\IEEEpeerreviewmaketitle

\IEEEraisesectionheading{\section{Introduction}\label{sec:introduction}}
\IEEEPARstart{T}{he} {ability to anticipate what is going to happen in the near future is fundamental for human beings in order to interact with the environment and make decisions.
Anticipation abilities are also fundamental to deploy intelligent systems which need to interact with a complex environment or other humans to automate challenging tasks and provide assistance.
Examples of such applications include autonomous vehicles~\cite{ma2019trafficpredict}, human-robotic symbiotic systems~\cite{koppula2016anticipating}, and wearable assistants~\cite{soran2015generating,kanade2012first}.
However, designing computational approaches to address tasks such as early action recognition~\cite{aliakbarian2017encouraging,de2016online,ma2016learning} and action anticipation~\cite{gao2017red,koppula2016anticipating,vondrick2016anticipating} is challenging as it often requires to model the relationship between past and future events, in the presence of incomplete observations.
First-Person (Egocentric) Vision~\cite{kanade2012first} offers an interesting scenario to study tasks related to anticipation.
On one hand, wearable cameras provide a means to collect naturally long videos containing multiple subsequent interactions with objects, which makes anticipation tasks unconstrained.
On the other hand, the ability to predict in advance what actions the camera wearer is going to perform and what objects they are going to interact with is useful to build intelligent wearable systems capable of anticipating the user's goals to provide assistance~\cite{kanade2012first}.}

\begin{figure}
	\includegraphics[width=\linewidth]{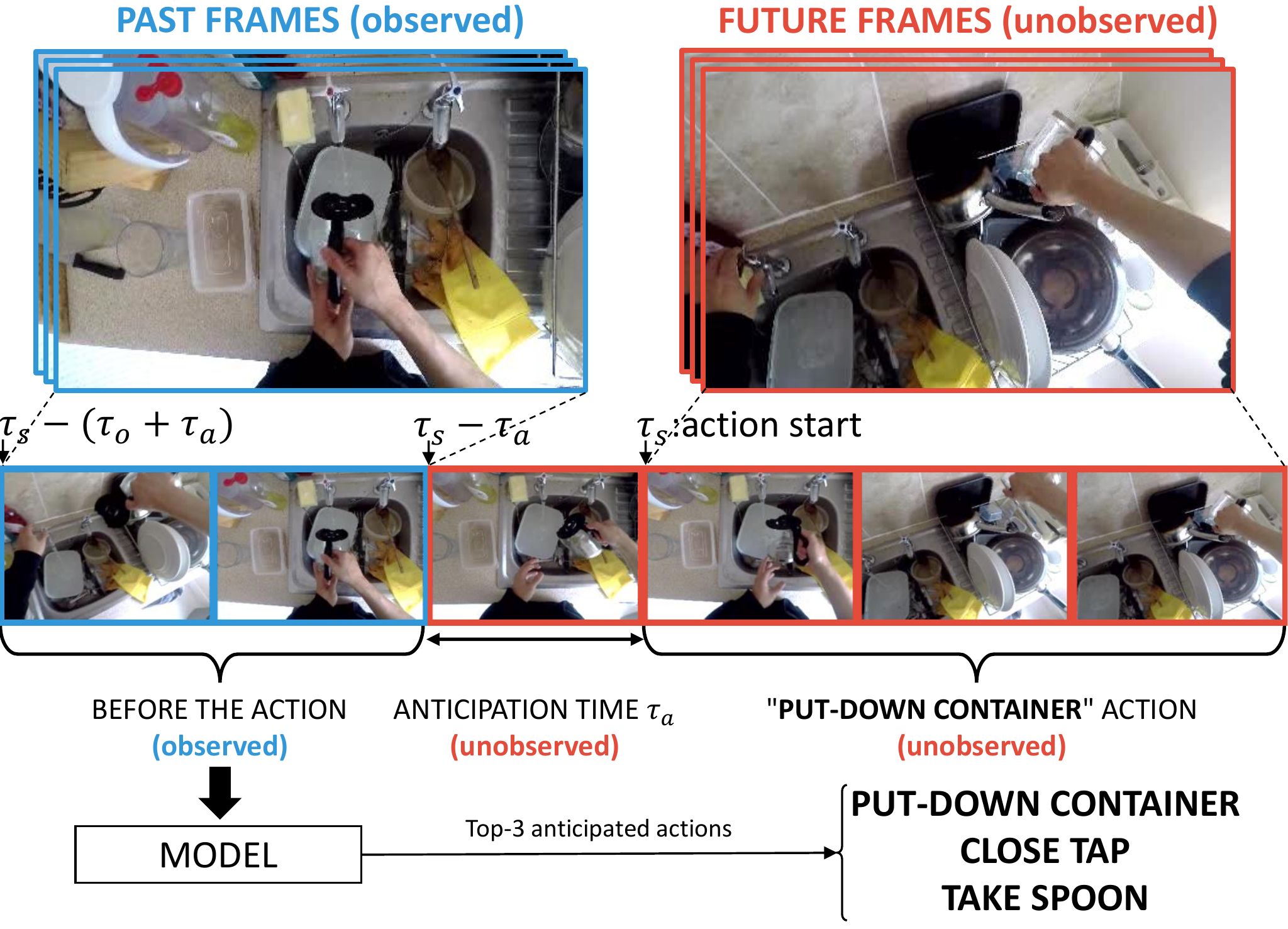}
	\caption{Egocentric Action Anticipation. See text for notation.}
	\label{fig:anticipation}
\end{figure}

{In this paper, we tackle the problem of egocentric action anticipation.
The task consists in recognizing a future action from an observation of the past. \figurename~\ref{fig:anticipation} illustrates the problem as defined in~\cite{Damen2018EPICKITCHENS}: given an action starting at time $\tau_s$, the system should predict the related action class by observing a video segment of temporal bounds $[\tau_s-(\tau_o+\tau_a), \tau_s-\tau_a]$, where $\tau_o$ denotes the ``observation time'', i.e. the length of the observed video, and $\tau_a$ denotes the ``anticipation time'', i.e., how much in advance the action has to be anticipated. Since the future is naturally uncertain, action anticipation models usually predict more than one possible action and the evaluation is performed using Top-k measures~\cite{koppula2016anticipating,furnari2018Leveraging}.
While the task of action anticipation has been investigated in the domain of third person vision~\cite{abu2018will,gao2017red,huang2014action,koppula2016anticipating,vondrick2016anticipating,jain2016recurrent}, less attention has been devoted to the challenging scenario of  egocentric videos~\cite{Damen2018EPICKITCHENS,furnari2018Leveraging,Ryoo2015a}.}

{Our work stems from the observation that egocentric action anticipation methods need to address two sub-tasks, which we refer to as ``encoding'' and ``inference''. In the encoding stage, the model has to summarize what has been observed in the past (e.g., ``a container has been washed'' in the observed segment in \figurename~\ref{fig:anticipation}). In the inference stage, the algorithm makes hypotheses about what may happen in the future, given the summary of the past and the current observation (e.g., ``put-down container'', ``close tap'', ``take spoon'' in \figurename~\ref{fig:anticipation}). 
Previous approaches have generally addressed the two sub-tasks jointly~\cite{aliakbarian2017encouraging,Damen2018EPICKITCHENS,ma2016learning,vondrick2016anticipating}. 
Our method is designed to disentangle them by using two separate LSTMs. ``A Rolling'' LSTM (R-LSTM) continuously encodes streaming observations and keeps an updated summary of what has been observed so far. When an anticipation is required, the ``Unrolling'' LSTM (U-LSTM) is initialized with the current hidden and cell states of the R-LSTM (which encode the summary of the past) and makes predictions about the future. 
While previous approaches considered fixed anticipation times~\cite{Damen2018EPICKITCHENS,furnari2018Leveraging,vondrick2016anticipating}, our architecture is designed to anticipate an action at multiple anticipation times. 
For instance, our model can anticipate actions from $2s$, to $0.25s$ before they occur, with the prediction refined as we get closer to the beginning of the action. 
To encourage the disentanglement of encoding and inference, we propose to pre-train our model with a novel ``Sequence Completion Pre-training'' (SCP) technique. 
Our method processes video in a multi-modal fashion, analyzing spatial observations (RGB frames), motion (optical flow) and object-based features obtained through an object detector.
We find that classic multimodal fusion techniques such as late and early fusion are limited in the context of action anticipation.
Therefore, we propose a novel ``Modality ATTention'' (MATT) mechanism to adaptively estimate optimal fusion weights for each modality by considering the outputs of the modality-specific R-LSTM components. 
We perform experiments on two large-scale datasets of egocentric videos, EPIC-KTICHENS~\cite{Damen2018EPICKITCHENS} and EGTEA Gaze+~\cite{Li_2018_ECCV}, and a standard benchmark of third person videos, Activitynet~\cite{caba2015activitynet}. 
the experiments show that the proposed method outperforms several state-of-the-art approaches and baselines in the task of egocentric action anticipation and generalizes to the scenario of third person vision, as well as to the tasks of early action recognition and action recognition. 
The proposed approach also achieved top performances in the 2019 EPIC-Kitchens egocentric action anticipation challenge\footnote{See \textit{https://epic-kitchens.github.io/Reports/EPIC-Kitchens-Challenges-2019-Report.pdf} for more details.}}

{The contributions of our work are the following: 1) we systematically investigate the problem of egocentric action anticipation within the framework provided by the EPIC-Kitchens dataset and its related challenges; 2) we benchmark popular ideas and approaches to action anticipation and define the proposed ``Rolling-Unrolling LSTM'' (RU) architecture, which is able to anticipate egocentric actions at multiple temporal scales; 3) we introduce two novel techniques specific to the investigated problem, i.e., i) ``Sequence Completion Pre-training'' and ii) adaptive fusion of multi-modal predictions with Modality ATTention (MATT); 4) we performed extensive evaluations to highlight the limits of previous methods and report improvements of the proposed approach over the state-of-the-art. To support future research in this field, we publicly release the code of our approach.}

{The reminder of this paper is organized as follows. Section~\ref{sec:related} revises the related works. Section~\ref{sec:method} details the proposed approach.
Section~\ref{sec:experiments} reports the experimental settings, whereas Section~\ref{sec:results} discusses the results.
Finally, Section~\ref{sec:conclusion} concludes the paper.}

{\section{Related Work}
\label{sec:related}
Our work is related to past research on action recognition, early action recognition, and anticipation tasks in both third and first-person vision.}

{\subsection{Action Recognition} 
Classic approaches to action recognition from video have generally relied on the extraction and processing of hand-designed features. Among the most notable approaches, Laptev~\cite{laptev2005space} proposed space-time interest points to classify events. Laptev et al.~\cite{laptev2008learning} further investigated the use of space-time features, space-time pyramids and SVMs for human action classification. Later, Wang et al.~\cite{wang2013dense,wang2013action} introduced dense trajectories to encode local motion and object appearance. More recent approaches investigated the use of deep learning to learn representations suitable to recognize actions directly from video. Karpathy et al.~\cite{karpathy2014large} considered the use of Convolutional Neural Networks (CNNs) and investigated different strategies to fuse per-frame predictions. Simonyan et al.~\cite{simonyan2014two} proposed Two-Stream CNN (2SCNN), a multi-branch architecture which recognizes actions by processing both appearance (RGB) and motion (optical flow) data. Feichtenhofer et al.~\cite{feichtenhofer2016spatiotemporal,feichtenhofer2017spatiotemporal,feichtenhofer2016convolutional} studied approaches to fuse predictions performed by the motion and appearance streams of a 2SCNN to improve recognition. Wang et al.~\cite{wang2016temporal} designed Temporal Segment Network (TSN), a general framework to train two-stream CNNs for action recognition. Zhou et al.~\cite{zhou2018temporal} introduced Temporal Relation Network (TRN), a module capable of encoding temporal dependencies between video frames at multiple time scales. Lin et al.~\cite{lin2018temporal} proposed the Temporal Shift Module (TSM), a component which facilitates information exchange among neighboring frames without the introduction of extra parameters in the network. 
Other authors investigated the use of 3D CNNs as a natural extension of 2D convolutional networks for video processing. Tran et al.~\cite{tran2015learning} demonstrated the use of 3D CNNs to learn spatio-temporal features for video classification. Carreira and Zisserman~\cite{carreira2017quo} proposed Inflated 3D (I3D) CNNs and showed how the weights of this architecture can be bootstrapped from a 2D CNN pre-trained on Imagenet. Hara et al.~\cite{hara2018can} studied whether 3D CNNs based on standard 2D ResNet~\cite{he2016deep} architectures could be exploited for action recognition from video. Tran et al.~\cite{tran2018closer} proposed R(2+1)D CNNs which factorize 3D convolutions as sequences of spatial and temporal convolutions. }

{Egocentric action recognition has also been studied in past works. Spriggs et al.~\cite{spriggs2009temporal} investigated the problem of supervised and unsupervised action segmentation using Inertial Measurement Units (IMU) and egocentric video. Fathi et al.~\cite{fathi2011understanding} proposed to recognize actions by modeling activities, hands and objects. Fathi et al.~\cite{fathi2012learning} employed eye gaze measurements to recognize egocentric actions. Pirsiavash and Ramanan~\cite{pirsiavash2012detecting} proposed to recognize egocentric activities using object-centric representations. Li et al.~\cite{li2015delving} studied how different egocentric cues, (including gaze, the presence of hands and objects, as well as head motion), can be used to perform the task. Ryoo et al.~\cite{ryoo2015pooled} proposed an approach to temporally pool features for egocentric action recognition. Ma et al.~\cite{ma2016going} designed a deep learning architecture which allows to integrate different egocentric-based features to recognize actions. Singh et al.~\cite{singh2017trajectory} adapted improved dense trajectories to the problem of egocentric action recognition. Singh et al.~\cite{singh2016first} proposed a multi-stream CNN to recognize egocentric actions using spatial features, temporal features and egocentric cues. Li et al.~\cite{Li_2018_ECCV} introduced a deep learning model for joint gaze estimation and action recognition in egocentric video. Sudhakaran et al.~\cite{sudhakaran2018lsta,sudhakaran2018attention} proposed to use a convolutional LSTM to recognize actions from egocentric video with an attention mechanism which learns to focus on image regions containing objects.}

{Our work builds on previous ideas investigated in the context of action recognition such as the use of multiple modalities for video analysis~\cite{simonyan2014two}, the use of Temporal Segment Networks~\cite{wang2016temporal} as a principled way to train CNNs for action recognition, as well as the explicit encoding of object-based features~\cite{fathi2011understanding,ma2016going,pirsiavash2012detecting,singh2016first,sudhakaran2018attention} to analyze egocentric video. However, in contrast with the aforementioned works, we address the problem of egocentric action \textit{anticipation} and show that approaches designed for action recognition, such as TSN~\cite{wang2016temporal} and late fusion to merge spatial and temporal predictions~\cite{simonyan2014two} are not directly applicable to the problem of egocentric action anticipation.}

{\subsection{Early Action Recognition in Third Person Vision}
Early action recognition refers to the task of recognizing an ongoing action as early as possible from partial observations~\cite{de2016online}. The problem of early action recognition has been widely investigated in the domain of third person vision. Ryoo~\cite{ryoo2011human} introduced the problem of recognizing ongoing actions from streaming video and addressed it proposing an integral histogram of spatio-temporal features. Cao et al.~\cite{cao2013recognize} used sparse coding to recognize actions from partially observed videos. Haoi and De la Torre~\cite{hoai2014max} proposed to use Structured Output SVM to detect partially observed events. Huang et al.~\cite{huang2014sequential} introduced Sequential Max-Margin Event Detectors, a method which performs early action detection by sequentially discarding classes until one class is identified as the detected one. De Geest et al.~\cite{de2016online} released a new dataset for online action detection and benchmarked several baseline methods to address the task. Ma et al.~\cite{ma2016learning} used LSTMs to address the problem of early action detection from video. Aliakbarian et al.~\cite{aliakbarian2017encouraging} proposed a two stage LSTM architecture which models context and action to perform early action recognition. Beccattini et al.~\cite{becattini2017done} designed ProgressNet, an approach capable of estimating the progress of actions and localizing them in space and time. De Geest and Tuytelaars~\cite{de2018modeling} addressed early action recognition proposing a ``feedback network'' which uses two LSTM streams to interpret feature representations and model the temporal structure of subsequent observations.}

{Differently from these works, we address the task of anticipating actions from egocentric video, i.e., predicting an action before it starts, and hence before it can be even partially observed. However, given the similarity between early action recognition and action anticipation, we consider and evaluate some ideas investigated in the context of early action recognition, such as the use of LSTMs to process streaming observations~\cite{aliakbarian2017encouraging,de2018modeling,ma2016learning} and the use of dedicated loss functions~\cite{ma2016learning}. Moreover, we show that the proposed architecture also generalizes to the problem of early egocentric action recognition, achieving state-of-the-art performance.}

{\subsection{Action Anticipation in Third Person Vision}
Action anticipation refers to the task of predicting an action \emph{before} it actually begins~\cite{gao2017red}. Previous works investigated different forms of action and activity anticipation from third person video. Kitani et al.~\cite{kitani2012activity} considered the task of inferring future paths followed by people observed from a static camera. Huang and Kitani~\cite{huang2014action} explored the task of action anticipation in the context of dual-agent interactions, where the actions of an agent are used to predict the response of the other agent. Lan et al.~\cite{lan2014hierarchical} proposed a hierarchical representation to anticipate future human actions from a still image or a short video clip. Jain et al.~\cite{jain2015car} designed an Autoregressive Input-Output HMM to anticipate driving maneuvers a few seconds before they occur using video, vehicle dynamics, GPS, and street maps. Jain et al.~\cite{jain2016recurrent} proposed a learning architecture based on LSTMs for driver activity anticipation. Koppula and Saxena~\cite{koppula2016anticipating} used object affordances to anticipate the possible future actions performed by a user from a robotic point of view. Vondrick et al.~\cite{vondrick2016anticipating} addressed action anticipation by training a CNN to regress the representations of future frames from past ones in an unsupervised way. Gao et al.~\cite{gao2017red} proposed an Encoder-Decoder LSTM architecture which predicts future actions by encoding the representations of past frames and regressing the representations of future frames. Similarly to~\cite{vondrick2016anticipating}, the model can be pre-trained from unlabeled videos in an unsupervised way. Felsen et al.~\cite{felsen2017will} developed a framework to forecast future events in team sports video from visual input. Mahmud et al.~\cite{mahmud2017joint} designed a system able to infer the labels and starting frames of future actions. Zeng et al.~\cite{zeng2017visual} introduced a general framework which uses inverse reinforcement learning to perform visual forecasting at different levels of abstraction, including story-line forecasting, action anticipation and future frames generation. Abu et al.~\cite{abu2018will} explored the use of CNNs and RNNs to predict the occurrence of future actions based on past observations.}

{In this work, we consider the problem of action anticipation from egocentric visual data. Nevertheless, our work builds on some of the ideas explored in past works such as the use of LSTMs~\cite{abu2018will,gao2017red,jain2016recurrent} to anticipate actions, the use of the encoder-decoder framework to encode past observations and produce hypotheses of future actions~\cite{gao2017red}, and the use of object specific features~\cite{mahmud2017joint} to determine which objects are present in the scene, we show that other approaches, such as the direct regression of future representations~\cite{gao2017red,vondrick2016anticipating}, do not achieve satisfactory performance in the egocentric scenario.}

{\subsection{Anticipation in First-Person Vision}
Past works have investigated different problems related to anticipation from first-person vision. Zhou and Berg~\cite{Zhou2015} studied methods to infer the ordering of egocentric video segments. Ryoo et al.~\cite{Ryoo2015a} proposed to analyze onset actions to anticipate potentially dangerous actions performed by humans against a robot. Soran et al.~\cite{soran2015generating} developed a system capable of inferring the next action performed in a known workflow to notify the user if a missing action is detected. Park et al.~\cite{SooPark2016} proposed a method to predict the future trajectories of the camera wearer from egocentric video. Zhang et al.~\cite{zhang2017deep} developed a method to predict eye gaze fixations in future video frames. Furnari et al.~\cite{Furnari2017} proposed to anticipate human-object interactions by analyzing the motion of objects in egocentric video. Chenyou et al.~\cite{chenyou2017forecasting} designed a method capable of forecasting the position of hands and objects in future frames. Rhinehart and Kitani~\cite{rhinehart2017first} used inverse reinforcement learning to anticipate future locations, objects and activities from egocentric video.}

{Previous works on anticipation from egocentric video have investigated different tasks and evaluated methods on different datasets and under different evaluation frameworks. Differently from these works, we consider the egocentric action anticipation challenge recently proposed by Damen et al.~\cite{Damen2018EPICKITCHENS}. It should be noted that few works~\cite{furnari2018Leveraging} have tackled the problem so far. While directly comparing our approach with respect to most of the aforementioned approach is unfeasible due to the lack of a common framework, our method incorporates some ideas from past works, such as the analysis of past actions~\cite{Ryoo2015a} and the detection of the objects present in the scene to infer future actions~\cite{Furnari2017}.}

\begin{figure}
	\includegraphics[width=\linewidth]{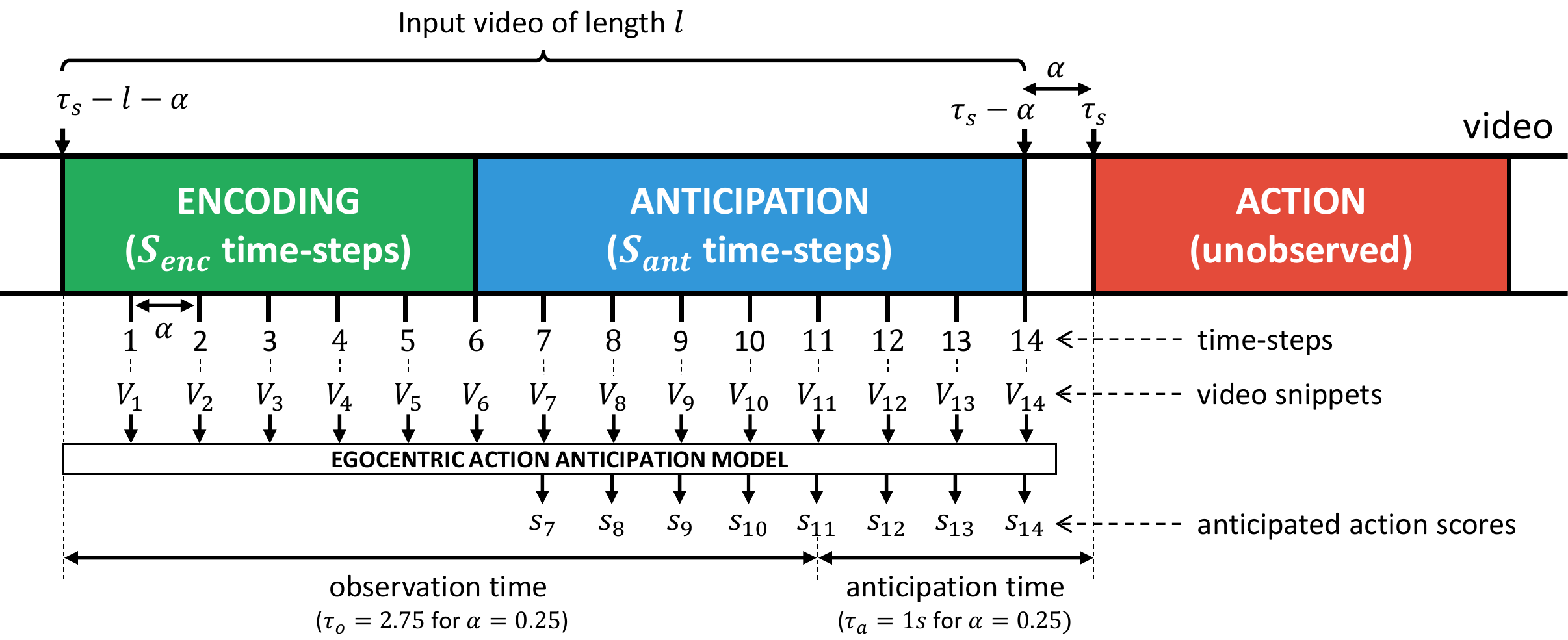}
	\caption{Video processing scheme adopted by the proposed method. In the example above, we set $S_{enc}=6$ and $S_{ant}=8$.}
	\label{fig:processing}
\end{figure}
\section{Proposed Approach}
\label{sec:method}
{In this Section, we discuss the proposed approach. Specifically, Section~\ref{sec:processing_scheme} introduces the strategy used to process video, Section~\ref{sec:rolling_unrolling} presents the proposed Rolling-Unrolling LSTMs module, Section~\ref{sec:scp} discusses the Sequence Completion Pre-Training (SCP) approach used to encourage the rolling and unrolling LSTMs to focus on different sub-taks, Section~\ref{sec:matt} introduces the modality attention mechanism used to fuse multi-modal predictions, Section~\ref{sec:rep_fun} details the definition of the representation functions used in the different branches of our architecture.}

\subsection{Video Processing Scheme}
\label{sec:processing_scheme}
{Past approaches~\cite{Damen2018EPICKITCHENS,vondrick2016anticipating,furnari2018Leveraging} performed action anticipation considering a fixed anticipation time $\tau_a$, usually set to $1$ second. 
This has been usually achieved by training a classifier to predict the action happening $\tau_a$ seconds after the end of an observed video segment. 
Similarly to~\cite{gao2017red}, we propose to anticipate actions at different temporal scales by using recurrent models. 
The authors of~\cite{gao2017red} obtain this multi-scale anticipation by training the model with variable anticipation times and performing inference using a fixed anticipation time chosen from the ones used during training. 
Also, the approach proposed in~\cite{gao2017red} requires the model to consume all the observed video before anticipating actions, which results in the separation between the observation and anticipation stages.
We argue that it would be natural to allow the model to make predictions \textit{while} observing the video and possibly refine them as more frames are processed.
We hence propose the video processing strategy illustrated in~\figurename~\ref{fig:processing}.}
According to the proposed scheme, the video is processed sequentially, with a video snippet $V_t$ consumed every $\alpha$ seconds, where $t$ indexes the current time-step.
At each time-step $t$, the model processes an input video snippet $V_t$ and optionally outputs a set of scores $s_t$ for the anticipated actions. 
Since the video is processed sequentially, the prediction made at time-step $t$ depends only on observations processed at previous time-steps. 
Specifically, the video is processed in two stages: an ``encoding'' stage, carried out for $S_{enc}$ time-steps and an ``anticipation'' stage, carried out for $S_{ant}$ time-steps. During the ``encoding'' stage, the model only observes incoming video snippets $V_i$ and does not anticipate actions. During the ``anticipation'' stage, the model both observes the input video snippets $V_i$ and outputs action scores $s_i$ for the anticipated actions.
This scheme effectively allows to anticipate actions at different anticipation times. In particular, in our experiments we set $\alpha=0.25s$, $S_{enc}=6$ and $S_{ant}=8$. 
In these settings, the model will process videos of length $l=\alpha(S_{enc}+S_{ant})=3.5s$ and output $8$ predictions at the following anticipation times: $\tau_a \in \{2s, 1.75s, 1.5s, 1.25s, 1s, 0.75s, 0.5s, 0.25s\}$. At time step $t$, the effective observation time will be given by $\alpha \cdot t$. 
Therefore, the $8$ predictions will be performed at the following observation times: $\tau_o \in \{1.75s, 2s, 2.25s, 2.5s, 2.75s, 3s, 3.25s, 3.5s\}$. 
It should be noted that our formulation generalizes the one proposed in~\cite{Damen2018EPICKITCHENS}. 
For instance, at time-step $t=11$, our model will anticipate actions with an effective observation time equal to $\tau_o=\alpha \cdot t=2.75s$ and an anticipation time equal to $\tau_a=\alpha(S_{ant}+S_{enc}+1-t)=1s$.

\subsection{Proposed Rolling-Unrolling LSTMs}
\label{sec:rolling_unrolling}

\begin{figure}
	\centering
	\includegraphics[width=\linewidth]{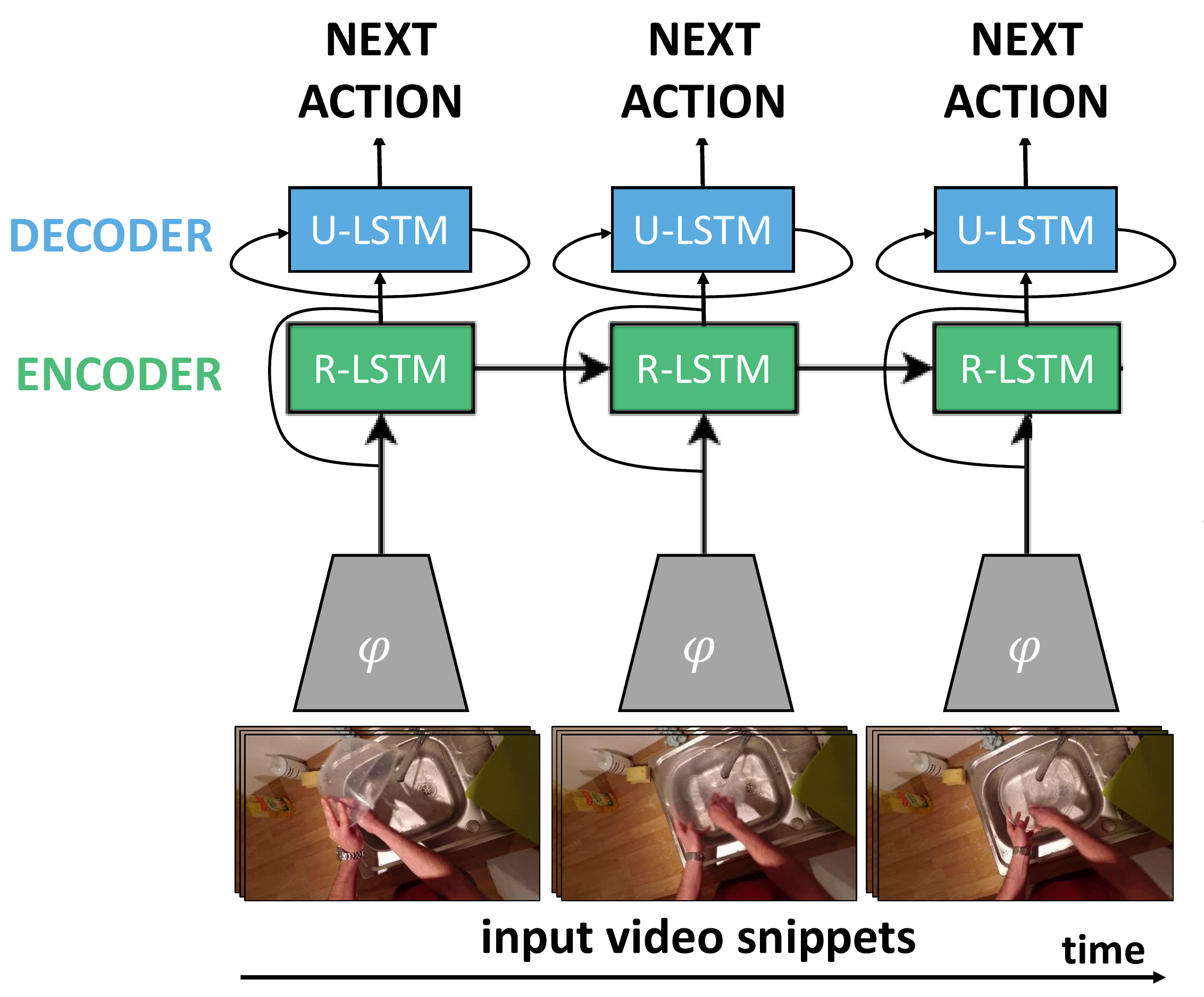}
	\caption{Overall architecture of the proposed Rolling-Unrolling LSTM architecture based on encoder-decoder models.}
	\label{fig:s2s}
\end{figure}

{Our model is inspired by encoder-decoder sequence to sequence models for text processing~\cite{sutskever2014sequence}.
Such models include an encoder which processes the words of the input sentence and a decoder which generates the words of the output sentence.
Both the encoder and decoder are often implemented using LSTMs.
Rather than analyzing words, our model processes high level representations of frames obtained through a representation function $\varphi$. 
The decoder is initialized with the internal representation of the encoder and iterates over the last representation to anticipate future actions. 
To allow for continuous anticipation of actions, the decoder is attached to each encoding time-step.
This allows to anticipate actions and refine predictions in a continuous fashion. 
We term the encoder ``Rolling LSTM'' (R-LSTM) and the decoder ``Unrolling LSTM'' (U-LSTM). Figure~\ref{fig:s2s} shows a diagram of the overall architecture of the proposed Rolling-Unrolling (RU) LSTMs, which is described in details below.}

Following previous literature~\cite{simonyan2014two}, we include multiple branches which analyze the video pre-processed according to different modalities.
Specifically, at each time-step $t$, the input video snippet $V_t$ is represented using different modality-specific representation functions $\varphi_1,\ldots,\varphi_M$, where $M$ is the number of considered modalities.
The representation functions can be learned and depend on the parameters $\theta^{\varphi_1},\ldots,\theta^{\varphi_M}$. 
This process allows to obtain modality-specific representations for the input video snippets $f_{1,t}=\varphi_1(V_t),\ldots,f_{M,t}=\varphi_M(V_t)$, where $f_{m,t}$ is the feature vector computed at time-step $t$ for modality $m$. 
The feature vector $f_{m,t}$ is hence fed to the $m^{th}$ branch of the architecture. In this work, we consider $M=3$ modalities, i.e., RGB frames (spatial branch), optical flow (motion branch) and object-based features (object branch).

\begin{figure}
g	\includegraphics[width=\linewidth]{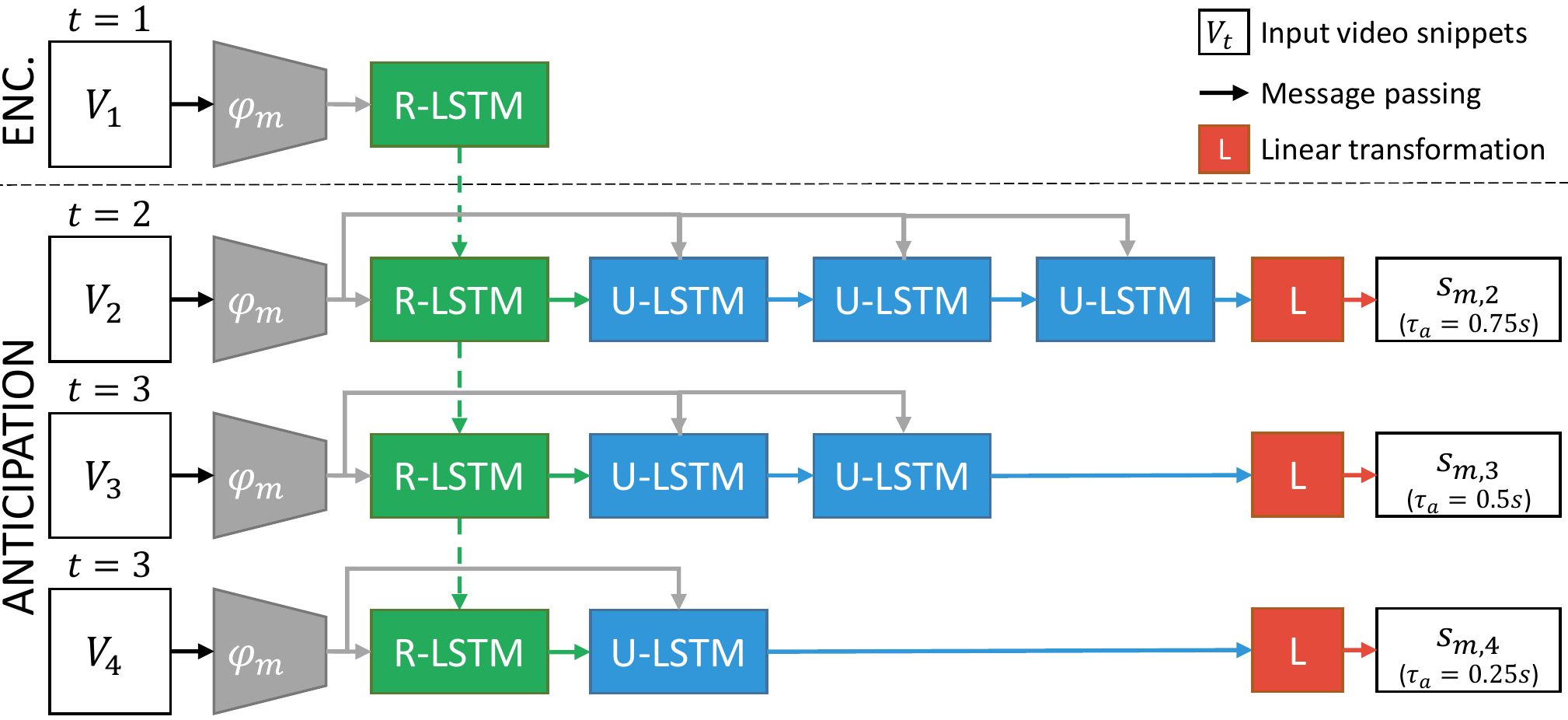}
	\caption{Example of modality-specific branch with $S_{enc}=1$ and $S_{ant}=3$.}
	\label{fig:branch}
\end{figure}

\figurename~\ref{fig:branch} illustrates in details the processing happening in a single branch $m$. For illustration purposes, the figure shows an example for $S_{enc}=1$ and $S_{ant}=3$. 
At a given time step $t$, the feature vector $f_{m,t}$ is fed to the R-LSTM, which is responsible for recursively encoding the semantic content of the incoming representations. 
This is performed according to the following equation:
\begin{equation}
(h_{m,t}^R,c_{m,t}^R)=LSTM_{\theta_m^R}(f_{m,t},h_{m,t-1}^R,c_{m,t-1}^R).
\end{equation}
In the equation above, $LSTM_{\theta_m^R}$ denotes the R-LSTM related to branch $m$, which depends on the learnable parameters $\theta_m^R$, whereas $h^R_{m,t}$ and $c^R_{m,t}$ are the hidden and cell states computed at time step $t$ in the branch related to modality $m$. The initial hidden and cell states of the R-LSTM are initialized with zeros: 
\begin{equation}
h^R_{m,0}=\textbf{0},\ c^R_{m,0}=\textbf{0}.
\end{equation}
Note that the $LSTM$ function follows the standard implementation of LSTMs~\cite{lstm1,lstm2}.

During the anticipation stage, at each time step $t$, the U-LSTM is used to predict future actions. The U-LSTM is initialized with the hidden and cell states of the R-LSTM at the current time-step: 
\begin{equation}
h_{m,0}^U=h_{m,t}^R,\ \ c_{m,0}^U=c_{m,t}^R
\end{equation} 
and iterates over the representation of the current video snippet $f_{m,t}$ for a number of times $n_t$ equal to the number of time-steps needed to reach the beginning of the action: $n_t=S_{ant}+S_{enc}+1-t$. Note that this number is proportional to the current anticipation time, which can be computed as $\alpha \cdot n_t$.
Similarly to the R-LSTM, the hidden and cell states of the U-LSTM are computed as follows at the generic iteration $j$:
\begin{equation}
(h_{m,j}^U,c_{m,j}^U)=LSTM_{\theta_m^U}(f_{m,t},h_{m,j-1}^U,c_{m,j-1}^U).
\label{eq:ulstm}
\end{equation}
In Equation~\eqref{eq:ulstm}, $LSTM_{\theta_m^U}$ represents the U-LSTM network related to branch $m$, which depends on the learnable parameters $\theta_m^U$. 
The vectors $h^U_{m,t}$ and $c^U_{m,t}$ are the hidden and cell states computed at iteration $j$ for modality $m$. 
It is worth noting that the input $f_{m,t}$ of the U-LSTM does not depend on $j$ (see Eq.~\eqref{eq:ulstm}), because it is fixed during the ``unrolling'' procedure. 
The main rationale of ``unrolling'' the U-LSTM for a different number of times at each time-step is to encourage the architecture to produce different predictions at different anticipation times.

Modality-specific anticipation scores $s_{m,t}$ for the anticipated actions are finally computed at time-step $t$ by feeding the last hidden vector $h^U_{m,n_t}$ of the U-LSTM to a linear transformation with learnable parameters $\theta^W_m$ and $\theta^b_m$: 
\begin{equation}
\label{eq:linear}
s_{m,t}=\theta^W_m h^U_{m,n_t}+\theta^b_m.
\end{equation}
{Anticipated action probabilities for modality $m$ at time-step $t$ are computed normalizing the scores $s_{m,t}$ with the Softmax function:
\begin{equation}
p_{m,t,i}=\frac{exp({s_{m,t,i}})}{\sum_k exp({s_{m,t,k}})}
\end{equation}
where $s_{m,t,i}$ denotes the $i^{th}$ component of the score vector $s_{m,t}$. A modality-specific RU branch is hence trained with the cross-entropy loss:
\begin{equation}
\label{eq:loss}
\mathcal{L}(p_{m},y)=-\frac{1}{S_{ant}}\sum_t \log p_{m,t,y}
\end{equation}
where $p_m$ is the set of probability distributions over actions computed by branch $m$ in all time-steps, $y$ is the ground truth class of the current sample and $\mathcal{L}$ is minimized with respect to the parameters $\theta_m^R$, $\theta_m^U$, $\theta_m^W$ and $\theta_m^b$.}

\subsection{Sequence Completion Pre-Training (SCP)}  
\label{sec:scp}
\begin{figure}
	\includegraphics[width=\linewidth]{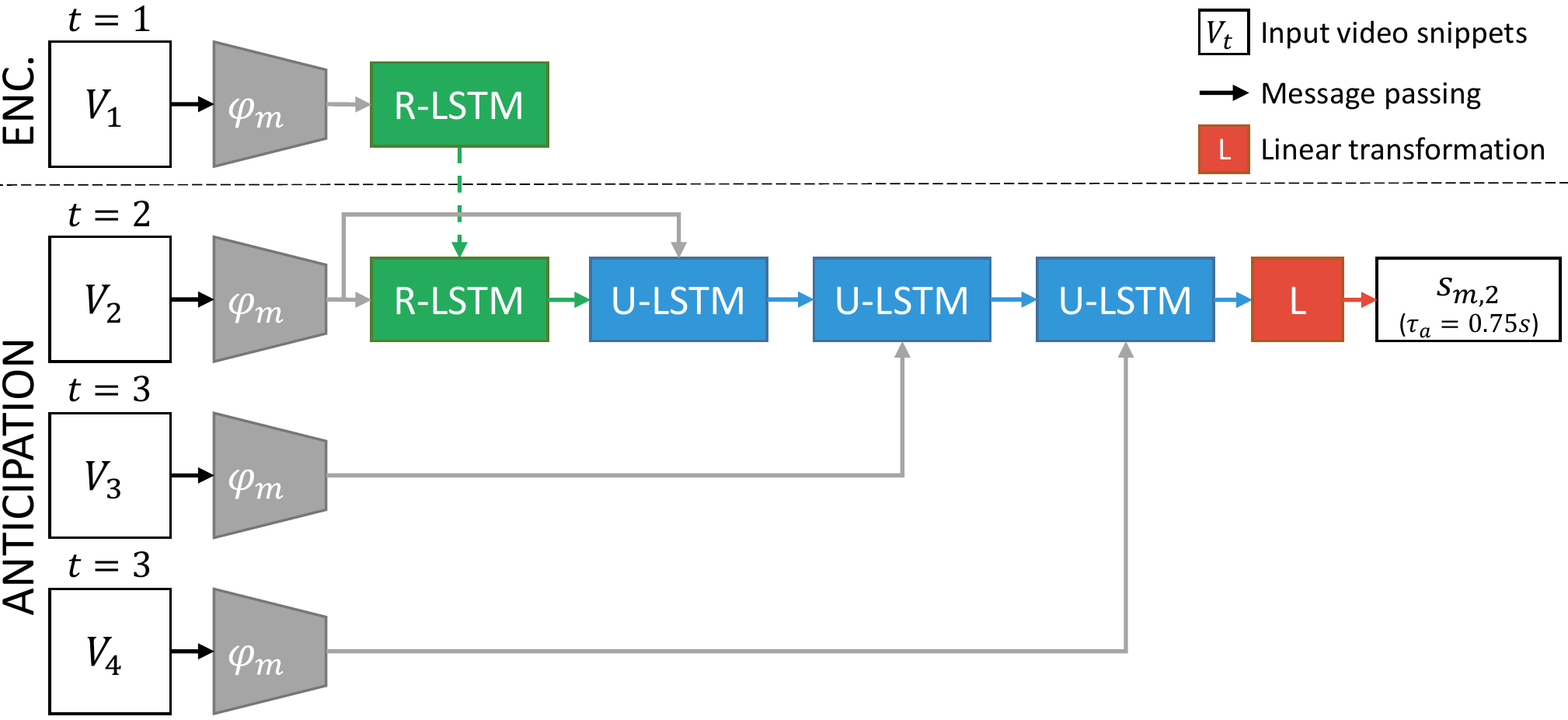}
	\caption{Example of connection scheme used during SCP for time-step $t=2$.}
	\label{fig:sequence_completion}
\end{figure}

The two LSTMs included in the proposed Rolling-Unrolling architecture are introduced to address two specific sub-tasks: the R-LSTM should encode past observations and summarize what has happened up to a given time-step, whereas the U-LSTM should focus on anticipating future actions conditioned on the hidden and cell vectors of the R-LSTM.
However, in practice, this might not happen. For instance, the R-LSTM could try to both summarize the past and anticipate future actions, which would make the task of the R-LSTM harder.
To encourage the two LSTMs to focus on the two different sub-tasks, we introduce a novel Sequence Completion Pre-training (SCP) procedure. 
During SCP, the connections of the network are modified to allow the U-LSTM to process future representations rather than iterating on the most recent one. 
In practice, the U-LSTM hidden and cell states are computed as follows during SCP:
\begin{equation}
\label{eq:scp}
(h_{m,j}^U,c_{m,j}^U)=LSTM_{\theta_m^U}(f_{m,t+j-1},h_{m,j-1}^U,c_{m,j-1}^U)
\end{equation}
where the input representations $f_{m,t+j-1}$ are sampled from future time-steps $t+j-1$. \figurename~\ref{fig:sequence_completion} illustrates an example of the connection scheme used during SCP for time-step $t=2$. Note that this is different from Eq.~\eqref{eq:ulstm}, in which only the most recent representation is processed. 
After SCP, the network is fine-tuned to the action anticipation task following Eq.~\eqref{eq:ulstm}.
The main goal of pre-training the model with SCP is to allow the R-LSTM to focus on summarizing past representations without trying to anticipate the future. 
Indeed, since the U-LSTM can ``cheat'' by looking into the future, the R-LSTM does not need to try to anticipate future actions to minimize the loss and is hence encouraged to focus on encoding.

\subsection{Modality ATTention (MATT)}
\label{sec:matt}
{Equation~\eqref{eq:linear} allows to obtain modality-specific action scores $s_{m,t}$ from the hidden representations of the U-LSTMs contained in each branch.
One way to fuse these scores is to compute a linear combination with a set of fixed weights $w_1,\ldots,w_M$, which is generally referred to as late fusion:
\begin{equation}
\label{eq:late_fusion}
s_t=w_1\cdot s_{1,t}+\ldots+w_M \cdot s_{M,t}.
\end{equation}
The fusion weights $w_m$ are fixed and generally found using cross validation.
We observe that, in the case of action anticipation, the relative importance of each modality might depend on the observed video.
For instance, in some cases the object detector computing object-based features might fail and hence become unreliable, or in other cases there could be little motion in the scenes, which would make the optical flow modality less useful.}
Inspired by previous work on attention~\cite{bahdanau2014neural,xu2015show} and multi-modal fusion~\cite{mees16iros}, we introduce a Modality ATTention (MATT) module which computes a set of attention scores indicating the relative importance of each modality for the final prediction. 
At a given time-step $t$, the attention scores are computed by feeding the concatenation of the hidden and cell vectors of the modality specific R-LSTM networks to a feed-forward neural network $D$ which depends on the learnable parameters $\theta^{MATT}$.
This computation is defined as follows:
\begin{equation}
\label{eq:matt}
\lambda_t=D_{\theta^{MATT}}(\oplus_{m=1}^M(h_{m,t}^R \oplus c_{m,t}^R))
\end{equation}
where $\oplus$ denotes the concatenation operator and $\oplus_{m=1}^M(h_{m,t}^R \oplus c_{m,t}^U)$ is the concatenation of the hidden and cell vectors produced by the R-LSTM at time-step $t$ across all modalities. 
Late fusion weights can be obtained by normalizing the score vector $\lambda_t$ using the softmax function, which makes sure that the computed fusion weights sum to one: 
\begin{equation}
\label{eq:matt_softmax}
w_{m,t}=\frac{e^{\lambda_{t,m}}}{\sum_k e^{\lambda_{t,k}}}
\end{equation}
where $\lambda_{t,m}$ is the $m^{th}$ component of the score vector $\lambda_t$. 
The final anticipated action scores are obtained at time-step $t$ by fusing the modality-specific predictions produced by the different branches with a linear combination as follows: 
\begin{equation}
\label{eq:matt_comb}
s_t=\sum_m w_{m,t} \cdot s_{m,t}.
\end{equation}%
\begin{figure}
	\includegraphics[width=\linewidth]{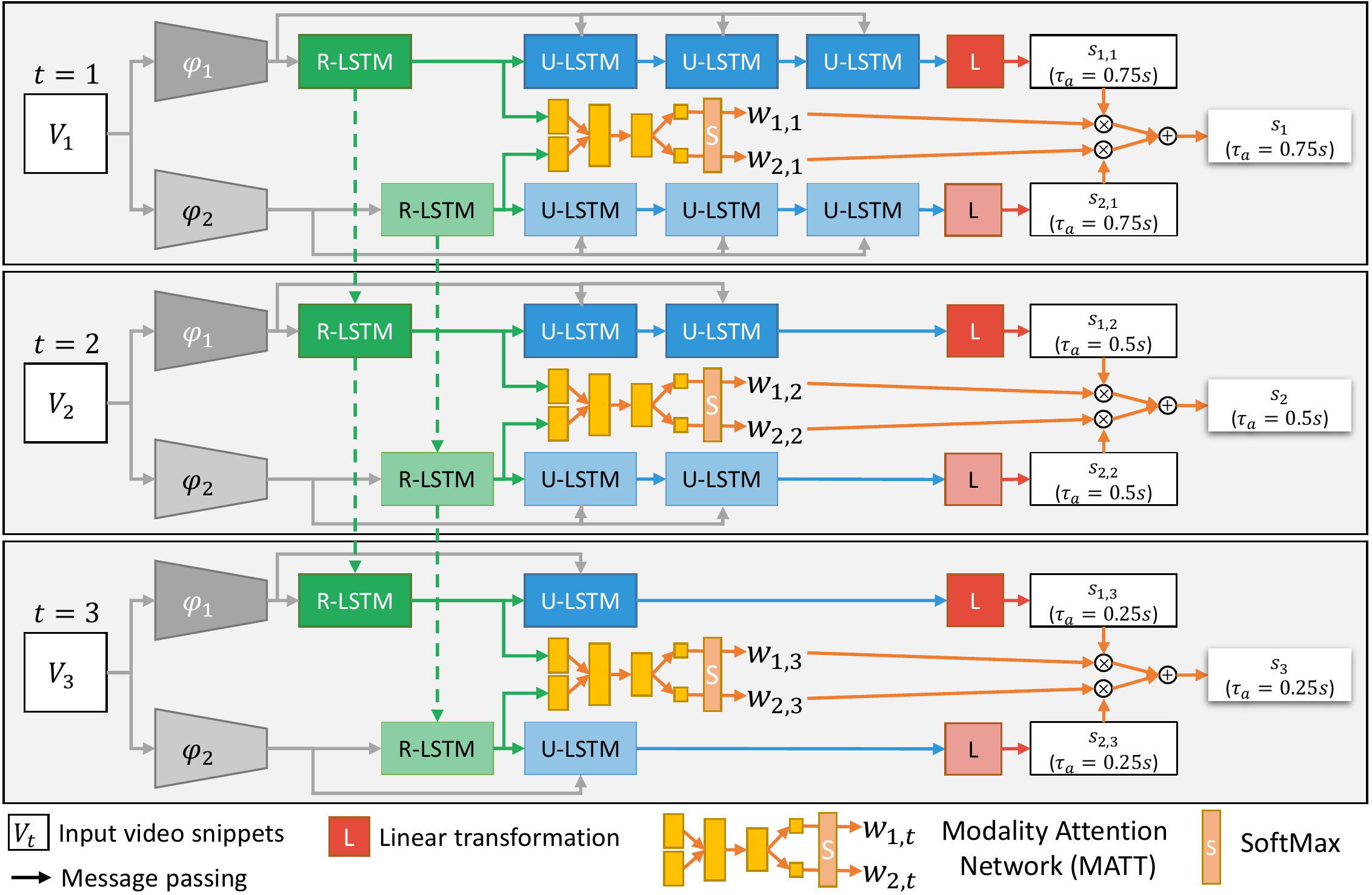}
	\caption{Example of the complete architecture with two branches and the Modality ATTention mechanism (MATT).}
	\label{fig:matt}
\end{figure}%
\figurename~\ref{fig:matt} illustrates an example of a complete RU with two modalities and the MATT fusion mechanism. For illustration purposes, the figure shows only three anticipation steps.
Note that, since the result of the linear combination defined in Eq.~\eqref{eq:matt_comb} is differentiable with respect to both the weights and the scores, the whole architecture is trainable end-to-end using the loss reported in Equation~\eqref{eq:loss}.

\subsection{Branches and Representation Functions}
\label{sec:rep_fun}
We instantiate the proposed architecture with $3$ branches: a spatial branch which processes RGB frames, a motion branch which processes optical flow fields, and an object branch which processes object-based features. 
The input to the representation functions are video snippets of $6$ frames $V_t=\{I_{t,1},I_{t,2},\ldots,I_{t,6}\}$, where $I_{t,i}$ is the $i^{th}$ frame of the video snippet $V_t$. 
The representation function $\varphi_1$ of the spatial branch is implemented as a Batch Normalized Inception CNN~\cite{ioffe2015batch} $CNN_{RGB}$ which is trained for action anticipation on the target dataset within the TSN framework~\cite{wang2016temporal}. {The CNN takes as input the last frame of each video snippet and extracts features from the last layer of $1024$ units preceding the final classifier. Hence:
\begin{equation}
\phi_1(V_t)=CNN_{RGB}(I_{t,6})
\end{equation}}

Similarly, the representation function $\varphi_2$ of the motion branch is implemented as a Batch Normalized Inception CNN~\cite{ioffe2015batch} $CNN_{Flow}$ pre-trained for action recognition on the target dataset following TSN~\cite{wang2016temporal}. The network analyzes a stack of optical flow fields computed from the $6$ frames ($5$ frame pairs) of the current video snippet as proposed in~\cite{wang2016temporal}. Similarly to $CNN_{RGB}$, $CNN_{Flow}$ extracts $1024$-dimensional features from the last layer preceding the final classifier. {This representation function is hence defined as:
\begin{equation}
\phi_2(V_t)=CNN_{Flow}(flow(I_{t,1},\ldots,I_{t,6}))
\end{equation}
where $flow$ computes the optical flow fields of the input frames and returns a tensor of $10$ channels obtained by stacking $x$ and $y$ optical flow images computed from frame pairs.}
It is worth noting that, since the CNNs have been trained for the action recognition task, $\varphi_1$ and $\varphi_2$ allow to obtain ``action-centric'' representations of the input frames, which can be used by the R-LSTM to summarize what has happened in the past. 

The representation function $\varphi_3$ related to the object branch includes an object detector $OD$ which detects objects in the last frame $I_{t,6}$ of the input video snippet $V_t$. 
When object-level annotations are available, the detector $OD$ is trained on the target dataset.
A fixed-length ``bag of objects'' representation is hence obtained by accumulating the confidence scores of all bounding boxes predicted for each object category. 
{Let $\{(b_{t,i},b^c_{t,i},b^s_{t,i})\}=OD(I_{t,6})$ be the set of bounding boxes $b_{t,i}$ predicted by the object detector along with the corresponding classes $b^c_{t,i}$ and confidence scores $b_{t,i}^s$.
The $j^{th}$ component of the fixed-length representation vector is obtained by summing the confidence scores of all objects detected for class $j$, i.e.,
\begin{equation}
boo(OD(I_{t,6}))_j=\sum_i [b^c_{t,i}=j]b^s_{t,i}
\end{equation}
where $boo$ is the ``bag of objects'' function and $[\cdot]$ denotes the Iverson bracket. The representation function can hence be defined as follows:
\begin{equation}
\phi_3(V_t)=boo(OD(I_{t,6}))
\end{equation}}
This representation encodes only the presence of an object in the scene, discarding its position in the frame, similarly to the representation proposed in~\cite{pirsiavash2012detecting} for egocentric activity recognition. 
We found this representation to be sufficient in the case of egocentric action anticipation. 
Differently from $\varphi_1$ and $\varphi_2$, $\varphi_3$ produces object-centric features which indicate what objects are likely to be present in the scene.\footnote{The reader is referred to Appendix~\ref{sec:implementation_details} for the implementation and training details of the proposed approach.}

\subsection{{Early Action Recognition and Action Recognition}}
{
We note that the proposed model can also be used for early action anticipation and action recognition. Specifically, this can be done by sampling a given number of frames $N$ from the video segment containing the action to be recognized and feeding the frames to the RU-LSTM model. To perform early action recognition, i.e., classifying the video before the action is completed, we set $S_{enc}=0$ and $S_{ant}=N$. The output of the model at the last time-step $t=N$ can be used to perform action recognition.}

\section{Experimental Settings}
\label{sec:experiments}
This section discusses the experimental settings, including the datasets considered for the evaluation, the evaluation measures and the compared methods.

\subsection{Datasets}
We performed experiments on two large-scale datasets of egocentric videos and a large-scale dataset of third person videos: EPIC-Kitchens~\cite{Damen2018EPICKITCHENS}, EGTEA Gaze+~\cite{Li_2018_ECCV} and ActivityNet~\cite{caba2015activitynet}. 
EPIC-Kitchens contains $39,596$ action annotations, $125$ verbs, and $352$ nouns. We split the public training set of EPIC-Kitchens ($28,472$ action segments) into training ($23,493$ segments) and validation ($4,979$ segments) sets by randomly choosing $232$ videos for training and $40$ videos for validation. 
We considered all unique $(verb, noun)$ pairs in the public training set, thus obtaining $2,513$ unique actions. 
EGTEA Gaze+ contains $10,325$ action annotations, $19$ verbs, $51$ nouns and $106$ unique actions. Methods are evaluated on EGTEA Gaze+ reporting the average performance across the three splits provided by the authors of the dataset~\cite{Li_2018_ECCV}.
{We considered the $1.3$ release of ActivityNet, which contains $10024$ training videos, $4926$ validation videos, and $5044$ test videos. Each video is labeled with one or more action segments belonging to one of $200$ action classes.
Videos are provided as YouTube links. Hence, depending on the country and time of download, some videos are not available. We have been able to download $7911$ training videos and $3832$ validation videos. 
Test videos are not used in our experiments as their labels are not publicly available.
The total number of training annotations amounts to $11890$, while the total amount of validation annotations is equal to $5786$. While the main focus of this work is on egocentric vision, we report experiments on this challenging dataset of third person videos to investigate the differences between the two scenarios and to what extent out approach generalizes to this domain.}

{We have extracted frames from the EPIC-Kitchens and EGTEA Gaze+ datasets using a constant frame-rate of $30fps$, whereas ActivityNet videos have been sub-sampled to $12fps$. All frames have been resized to $456 \times 256$ pixels. Optical flow fields have been computed on all datasets using the TVL1 algorithm~\cite{zach2007duality}.}

\begin{table*}
	\caption{Egocentric action anticipation results on the EPIC-KITCHENS dataset.}
	\label{tab:anticipation_ek}
	\begin{adjustbox}{width=\linewidth,center}
		\begin{tabular}{p{1.5cm}cccccccc||ccc|ccc|ccc}
			\hline
			\multicolumn{1}{c}{} & \multicolumn{8}{c||}{Top-5 ACTION Accuracy\% @ different $\tau_a$(s)} & \multicolumn{3}{c|}{Top-5 Acc.\% @1s} & \multicolumn{3}{c|}{M. Top-5 Rec.\% @1s} & \multicolumn{3}{c}{Mean $TtA(5)$}   \\  \hline %
			\multicolumn{1}{c}{} & $2$ & $1.75$ & $1.5$ & $1.25$ & $1.0$& $0.75$&$0.5$&$0.25$ & VERB & NOUN & ACT. & VERB & NOUN & ACT. & VERB & NOUN & ACT.\\ \hline
			DMR~\cite{vondrick2016anticipating} & /              & /              & /              & /              & 16.86          & /              & /              & /              & 73.66          & 29.99          & 16.86          & 24.50          & 20.89          & 03.23          & /              & /              & /              \\
			ATSN~\cite{Damen2018EPICKITCHENS}  & /              & /              & /              & /              & 16.29          & /              & /              & /              & \underline{77.30} & 39.93          & 16.29          & 33.08          & 32.77          & 07.60          & /              & /              & /              \\
			MCE~\cite{furnari2018Leveraging}  & /              & /              & /              & /              & 26.11          & /              & /              & /              & 73.35          & 38.86          & 26.11          & 34.62          & 32.59          & 06.50          & /              & /              & /              \\
			{TCN~\cite{BaiTCN2018}}  &19.33 & 19.95 & 20.43 & 20.82 & 21.82 & 23.03 & 23.35 & 24.40 & 73.93 & 36.75 & 21.82 & 28.95 & 30.28 & 05.28 & 01.54 & 00.88 & 00.56\\
			{ED*~\cite{gao2017red}}&21.45 & 22.37 & 23.26 & 24.51 & 25.20 & 26.34 & 27.45 & 28.67 & 76.24 & 42.18 & 25.20 & \underline{42.25} & 42.00 & 09.98 & 01.59 & 00.99 & 00.61\\
			ED~\cite{gao2017red}           & 21.53          & 22.22          & 23.20          & 24.78          & 25.75          & 26.69          & 27.66          & 29.74          & 75.46          & 42.96          & 25.75          & {41.77}          & \underline{42.59}          & \underline{10.97}          & \underline{01.60} & \underline{01.02} & \underline{00.63} \\
			FN~\cite{de2018modeling}       & 23.47          & 24.07          & 24.68          & 25.66          & 26.27          & 26.87          & 27.88          & 28.96          & 74.84          & 40.87          & 26.27          & 35.30          & 37.77          & 06.64          & 01.52          & 00.86          & 00.56          \\
			RL~\cite{ma2016learning} & 25.95 & 26.49 & 27.15 & 28.48 & 29.61 & 30.81 & 31.86 & 32.84 & 76.79 & \underline{44.53} & 29.61 & 40.80 & 40.87 & 10.64 & 01.57 & 00.94 & 00.62\\
			EL~\cite{jain2016recurrent}    & 24.68          & 25.68          & 26.41          & 27.35          & 28.56          & 30.27          & 31.50          & \underline{33.55} & 75.66          & 43.72          & 28.56          & 38.70          & 40.32          & 08.62          & 01.55          & 00.94          & 00.62          \\
			{LSTM~\cite{lstm2}}           & \underline{26.45} & \underline{27.11} & \underline{28.22} & \underline{29.24} & \underline{29.89} & \underline{31.03} & \underline{31.88} & {33.19} & 76.33 & 44.21 & \underline{29.89} & 39.31 & 40.30 & 10.42 & 01.56 & 00.93 & \underline{00.63}\\

			\textbf{RU-LSTM}                             & \textbf{29.44} & \textbf{30.73} & \textbf{32.24} & \textbf{33.41} & \textbf{35.32} & \textbf{36.34} & \textbf{37.37} & \textbf{38.98} & \textbf{79.55} & \textbf{51.79} & \textbf{35.32} & \textbf{43.72} & \textbf{49.90} & \textbf{15.10} & \textbf{01.62} & \textbf{01.11} & \textbf{00.76} \\ \hline
			Improvement                                 & +2.99          & +3.62          & +4.02          & +4.17          & +5.43          & +5.31          & +5.49          & +5.43          & +2.25          & +7.26          & +5.43          & +1.47          & +7.31          & +4.13          & +0.02          & +0.09          & +0.13          \\ \hline
		\end{tabular}
	\end{adjustbox}
	
\end{table*}

\subsection{Evaluation Measures}
We evaluate all methods using Top-k evaluation measures, i.e., we consider a prediction to be correct if the ground truth action label is included in the top-k predictions. 
As observed in previous works~\cite{furnari2018Leveraging,koppula2016anticipating}, this evaluation scheme is appropriate given the uncertainty of future predictions (i.e., many plausible actions can be performed after an observation).
Specifically, we use the Top-5 accuracy as a class-agnostic measure and the Mean Top-5 Recall as a class aware metric. The Top-5 recall for a given class $c$ is defined as the fraction of samples of ground truth class $c$ for which the class $c$ is in the list of the top-5 anticipated actions~\cite{furnari2018Leveraging}. 
The mean Top-5 Recall is obtained by averaging the Top-5 recall values over classes. 
When evaluating on EPIC-Kitchens, Top-5 Recalls are averaged over the provided list of many-shot verbs, nouns and actions. Results on the EPIC-Kitchens official test set are reported using the suggested evaluation measures, i.e., Top-1 accuracy, Top-5 accuracy, Precision and Recall. Early action recognition and action recognition models are evaluated using Top-1 accuracy.

To assess the timeliness of anticipations, we propose a novel evaluation measure inspired by the AMOC curve~\cite{hoai2014max}. Let $s_t$ be the action scores predicted at time-step $t$ for an action of ground truth class $c$. Let $\tau_t$ be the anticipation time at time-step $t$, and $tk(s_t)$ be the set of top-$k$ actions as ranked by the action scores $s_t$. We define as ``time to action'' at rank $k$ the largest anticipation time (i.e., the time of earliest anticipation) in which a correct prediction has been made according to the Top-$k$ criterion: 
\begin{equation}
TtA(k)=max\{\tau_t |c \in tk(s_t),\ \forall t\}
\end{equation}
If an action is not correctly anticipated in any of the time-steps, we set $TtA(k)=0$. 
The mean time to action over the whole test set $mTtA(k)$ indicates how early, in average, a method can anticipate actions.

{The time to action measure can be extended also to the case of early action recognition. If the current video is composed by $N$ frames, we define the observation ratio at time-step $t$ as $OR(t)=\frac{t}{N}$. This number can also be interpreted as a percentage, which defines how much of the action has been observed so far. 
We hence define as the ``Minimum Observation Ratio'' (MOR) the smallest observation ratio in which a correct prediction has been made according to the Top-1 criterion:
\begin{equation}
MOR=min\{OR(t) | c = argmax_{j}\{s_{t,j}\},\ \forall t\}.
\end{equation}}

We evaluated performances for verb, noun and action predictions on the EPIC-Kitchens and EGTEA Gaze+ datasets.
We obtained verb and noun scores by marginalization over the action scores for all methods except the one proposed in~\cite{Damen2018EPICKITCHENS}, which predicts verb and noun scores directly.

\subsection{Compared Methods}
We compare the proposed method with respect to several state-of-the approaches and baselines. 
Specifically, we consider the Deep Multimodal Regressor (DMR) proposed in~\cite{vondrick2016anticipating}, the Anticipation Temporal Segment Network (ATSN) of~\cite{Damen2018EPICKITCHENS}, the anticipation Temporal Segment Network trained with verb-noun Marginal Cross Entropy Loss (MCE) described in~\cite{furnari2018Leveraging}, and the Encoder-Decoder LSTM (ED) introduced in~\cite{gao2017red}.
{We also consider two standard sequence-to-sequence models: a single LSTM architecture~\cite{lstm1,lstm2} (LSTM), and Temporal Convolutional Networks~\cite{BaiTCN2018} (TCN).}
We further compare our approach with respect to the following methods originally proposed for the task of early action recognition: a single LSTM architecture (we use the same parameters as our R-LSTM) trained using the Ranking Loss on Detection Score proposed in~\cite{ma2016learning} (RL), an LSTM trained using the Exponential Anticipation Loss proposed in~\cite{jain2016recurrent} (EL), and the Feedback Network LSTM (FN) proposed in~\cite{de2018modeling}. 
Note that, being essentially sequence-to-sequence models, these approaches can be easily adapted to the considered action anticipation scenario.
All these baselines adopt the video processing scheme illustrated in \figurename~\ref{fig:processing}. Among them, LSTM, RL, FN and EL are implemented as two stream networks with a spatial and a temporal branch whose predictions are fused by late fusion.
{In our experiments, TCN obtained very low performance when processing optical flows on the EPIC-Kitchens and EGTEA Gaze+ datasets.
In these cases, fusing the RGB and Flow branches actually resulted in lower performances than the RGB branch alone.
On the contrary, on the ActivityNet dataset, fusing the RGB and Flow branches led to better performance. 
Hence, we implemented TCN as a single RGB branch on EPIC-Kitchens and EGTEA Gaze+ and as a two-branch network with late fusion on ActivityNet.\footnote{The reader is referred to Appendix~\ref{sec:implementation_details_compared} for the implementation details of the considered methods.}
Additionally, we compare our approach with Two-Stream CNNs (2SCNN)~\cite{simonyan2014two} and the method proposed by Miech et al~\cite{miech2019leveraging} on the official test sets of EPIC-Kitchens.}

\section{Results}
\label{sec:results}
{This section compares the performance of the proposed Rolling-Unrolling LSTMs with other state-of-the-art approaches. Specifically, Sections~\ref{sec:res_ek}-\ref{sec:res_an} discuss the action anticipation results on the three considered datasets, Section~\ref{sec:res_ablation} reports the ablation study on the EPIC-Kitchens dataset, whereas Section~\ref{sec:res_qualitative} reports some qualitative examples of the proposed method.}

\subsection{Egocentric Action Anticipation on EPIC-Kitchens}
\label{sec:res_ek}
\tablename~\ref{tab:anticipation_ek} compares RU with respect to the other state-of-the-art approaches on our validation set of the EPIC-Kitchens dataset. The left part of the table reports Top-5 action anticipation accuracy for the $8$ considered anticipation times. Note that some methods~\cite{Damen2018EPICKITCHENS,furnari2018Leveraging,vondrick2016anticipating} have been designed to anticipate actions only at a fixed anticipation time. 
The right part of the table reports the Top-5 accuracy and Mean Top-5 Recall for verbs, nouns and actions, for the fixed anticipation time of $\tau_a=1s$, as well as the mean $TtA(5)$ scores obtained across the validation set. 
Best results are highlighted in bold, whereas second-best results are underlined. The last row reports the improvements obtained by RU with respect to second-best results. {ED* denotes the Encoder-Decoder approach proposed in~\cite{gao2017red} without the unsupervised pre-training procedure proposed by the authors. These results are reported for reference.}

\begin{table*}[t]
	\caption{Egocentric action anticipation results on the EPIC-Kitchens test set.}
	\label{tab:anticipation_ek_test}
	\begin{adjustbox}{width=\linewidth,center}
		\setlength{\tabcolsep}{3pt}
		\begin{tabular}{llccc|ccc|ccc|ccc}
			& & \multicolumn{3}{c|}{Top-1 Accuracy\%} & \multicolumn{3}{c|}{Top-5 Accuracy\%} & \multicolumn{3}{c|}{Avg Class Precision\%} & \multicolumn{3}{c}{Avg Class Recall\%} \\ \hline
			& & VERB & NOUN & ACTION & VERB & NOUN & ACTION & VERB & NOUN & ACTION & VERB & NOUN & ACTION \\ \hline
			\multirow{7}{*}{\rotatebox{90}{\textbf{S1}}} &
			{DMR~\cite{vondrick2016anticipating}} & 26.53 & 10.43 & 01.27 & 73.30 & 28.86 & 07.17 & 06.13 & 04.67 & 00.33 & 05.22 & 05.59 & 00.47\\
			&
			{2SCNN~\cite{Damen2018EPICKITCHENS}} & 29.76 & 15.15 & 04.32 & 76.03 & 38.56 & 15.21 & 13.76 & 17.19 & 02.48 & 07.32 & 10.72 & 01.81\\
			&ATSN~\cite{Damen2018EPICKITCHENS} & \underline{31.81} & 16.22 & 06.00 & \underline{76.56} & 42.15 & \underline{28.21} & \underline{23.91} & \underline{19.13} & 03.13 & 09.33 & 11.93 & 02.39\\
			&MCE~\cite{furnari2018Leveraging} & 27.92 & 16.09 & \underline{10.76} & 73.59 & 39.32 & 25.28 & 23.43 & 17.53 & \underline{06.05} & \underline{14.79} & 11.65 & \underline{05.11}\\
			&{ED~\cite{gao2017red}} & 29.35 & 16.07 & 08.08 & 74.49 & 38.83 & 18.19 & 18.08 & 16.37 & 05.69 & 13.58 & \underline{14.62} & 04.33\\
			&{Miech et al.~\cite{miech2019leveraging}} & 30.74 & \underline{16.47} & 09.74 & 76.21 & \underline{42.72} & 25.44 & 12.42 & 16.67 & 03.67 & 08.80 & 12.66 & 03.85\\
			&RU-LSTM & 
			\textbf{33.04} & \textbf{22.78} & \textbf{14.39} & \textbf{79.55} & \textbf{50.95} & \textbf{33.73} & \textbf{25.50} & \textbf{24.12} & \textbf{07.37} & \textbf{15.73} & \textbf{19.81} & \textbf{07.66}\\ \hline
			&Imp. wrt best & +1.23 & +6.31 & +3.63 & +2.99 & +8.23 & +5.52 & +1.59 & +4.99 & +1.32 & +0.94 & +5.19 & +2.55\\

			\hline
			\multirow{7}{*}{\rotatebox{90}{\textbf{S2}}} &
			{DMR~\cite{vondrick2016anticipating}} & 24.79 & 08.12 & 00.55 & 64.76 & 20.19 & 04.39 & 09.18 & 01.15 & 00.55 & 05.39 & 04.03 & 00.20\\
			&{2SCNN~\cite{Damen2018EPICKITCHENS}} & 25.23 & 09.97 & 02.29 & {68.66} & 27.38 & 09.35 & \textbf{16.37} & 06.98 & 00.85 & 05.80 & 06.37 & 01.14\\
			&ATSN~\cite{Damen2018EPICKITCHENS} & {25.30} & {10.41} & 02.39 & 68.32 & {29.50} & 06.63 & 07.63 & \underline{08.79} & 00.80 & 06.06 & {06.74} & 01.07\\
			&MCE~\cite{furnari2018Leveraging} & 21.27 & 09.90 & {05.57} & 63.33 & 25.50 & {15.71} & 10.02 & 06.88 & {01.99} & {07.68} & 06.61 & {02.39}\\
			&{ED~\cite{gao2017red}} & 22.52 & 07.81 & 02.65 & 62.65 & 21.42 & 07.57 & 07.91 & 05.77 & 01.35 & 06.67 & 05.63 & 01.38\\
			&{Miech et al.\cite{miech2019leveraging}} & \textbf{28.37} & \underline{12.43} & \underline{07.24} & \textbf{69.96} & \underline{32.20} & \underline{19.29} & 11.62 & 08.36 & \underline{02.20} & \underline{07.80} & \underline{09.94} & \underline{03.36}\\

			&RU-LSTM & \underline{27.01} & \textbf{15.19} & \textbf{08.16} & \underline{69.55} & \textbf{34.38} & \textbf{21.10} & {\underline{13.69}} & \textbf{09.87} & \textbf{03.64} & \textbf{09.21} & \textbf{11.97} & \textbf{04.83}\\ \hline
			&Imp. wrt best & -1.36 & +2.76 & +0.92 & -0.41 & +2.18 & +1.81 & -2.68 & +1.08 & +1.44 & +1.41 & +2.03 & +1.47\\

			\hline

		\end{tabular}
	\end{adjustbox}	
\end{table*}

\begin{table*}
	\caption{Egocentric action anticipation results on EGTEA Gaze+.}
	\label{tab:anticipation_egtea}
	\begin{adjustbox}{width=\linewidth,center}
		\begin{tabular}{p{1.5cm}cccccccc||ccc|ccc|ccc}
			\hline
			\multicolumn{1}{c}{} & \multicolumn{8}{c||}{Top-5 ACTION Accuracy\% @ different $\tau_a$(s)} & \multicolumn{3}{c|}{Top-5 Acc.\% @1s} & \multicolumn{3}{c|}{M. Top-5 Rec.\% @1s} & \multicolumn{3}{c}{Mean $TtA(5)$}   \\  \hline %
			\multicolumn{1}{c}{} & $2$ & $1.75$ & $1.5$ & $1.25$ & $1.0$& $0.75$&$0.5$&$0.25$ & VERB & NOUN & ACT. & VERB & NOUN & ACT. & VERB & NOUN & ACT.\\ \hline
			DMR~\cite{vondrick2016anticipating} & / & / & / & / & 55.70 & / & / & / & \underline{92.78} & 71.36 & 55.70 & 70.22 & 53.92 & 38.11 & / & / & /\\
			ATSN~\cite{Damen2018EPICKITCHENS}  & / & / & / & / & 40.53 & / & / & / & 90.60 & 69.94 & 40.53 & 69.24 & 57.02 & 31.61 & / & / & /\\
			MCE~\cite{furnari2018Leveraging}  & / & / & / & / & 56.29 & / & / & / & 90.73 & 70.02 & 56.29 & 72.38 & 58.67 & 43.75 & / & / & /\\
			{TCN~\cite{BaiTCN2018}} & 49.86 & 51.05 & 54.08 & 55.17 & 58.50 & 59.34 & 62.87 & 65.53 & 91.10 & 71.94 & 58.50 & 73.36 & 63.11 & 47.14 & {01.86} & \underline{01.58} & {01.33}\\
			{ED*}~\cite{gao2017red}   & 52.91 & 54.16 & 56.22 & 58.31 & 60.18 & 62.57 & 64.77 & 67.05 & 91.12 & 73.50 & 60.18 & 78.19 & 68.33 & 54.61 & \underline{01.87} & 01.57 & \underline{01.34}\\        
			ED~\cite{gao2017red}           &  45.03 & 46.22 & 46.86 & 48.36 & 50.22 & 51.86 & 49.99 & 49.17 & 86.79 & 64.35 & 50.22 & 69.66 & 56.62 & 42.74 & 01.84 & 01.40 & 01.24\\
			FN~\cite{de2018modeling}       & 54.06 & 54.94 & 56.75 & 58.34 & 60.12 & 62.03 & 63.96 & 66.45 & 91.05 & 71.64 & 60.12 & 76.73 & 63.59 & 49.82 & 01.83 & 01.39 & 01.26\\
			RL~\cite{ma2016learning}       & 55.70 & 56.45 & 58.65 & 60.69 & 62.74 & 64.37 & 67.02 & 69.33 & 91.54 & 74.51 & 62.74 & 78.55 & 67.10 & 52.17 & 01.84 & 01.43 & 01.29\\
			EL~\cite{jain2016recurrent}    & 55.05 & {56.75} & {58.81} & 61.00 & {63.76} & \underline{66.37} & \underline{69.12} & \underline{72.33} & 91.77 & \underline{75.68} & {63.76} & \underline{79.63} & \underline{69.93} & \underline{55.11} & {01.85} & {01.47} & {01.32}\\
			{LSTM~\cite{lstm2}}           & \textbf{56.88} & \underline{58.23} & \underline{59.87} & \underline{61.83} & \underline{63.87} & 65.84 & 67.70 & 70.65 & 91.56 & 75.30 & \underline{63.87} & 78.27 & 68.43 & 53.35 & {01.85} & {01.54} & {01.33}\\ 
			\textbf{RU-LSTM}             & \underline{56.82} & \textbf{59.13} & \textbf{61.42} & \textbf{63.53} & \textbf{66.40} & \textbf{68.41} & \textbf{71.84} & \textbf{74.28} & \textbf{93.11} & \textbf{77.48} & \textbf{66.40} & \textbf{82.07} & \textbf{73.30} & \textbf{58.64} & \textbf{01.88} & \textbf{01.61} & \textbf{01.41}\\
			\hline
			Improv.                                 & -0.06 & +0.90 & +1.55 & +1.70 & +2.53 & +2.04 & +2.72 & +1.95 & +0.33 & +1.80 & +2.53 & +2.44 & +3.37 & +3.53 & +0.01 & +0.03 & +0.07\\
			\hline
		\end{tabular}
	\end{adjustbox}
	
\end{table*}

The proposed approach outperforms all competitors by consistent margins according to all evaluation measures, obtaining an average improvement over prior art of about $5\%$ with respect to Top-5 action anticipation accuracy on all anticipation times. 
The methods based on TSN (ATSN and MCE) tend to achieve low performance, which suggests the limits of simply adapting action recognition methods to the problem of anticipation. 
Interestingly, DMR and ED, which are explicitly trained to anticipate future representations, achieve sub-optimal Top-5 action anticipation accuracy as compared to methods trained to predict future actions directly from input images (e.g., compare DMR with MCE, and ED with FN/RL/EL/{LSTM/TCN}/RU).
{Comparing ED* to ED reveals that the unsupervised pre-training based on the regression of future representations is not beneficial in the considered problem of egocentric action anticipation. Indeed, in most cases the results achieved by the two methods are comparable.}
This might be due to the fact that anticipating future representations is very challenging in the case of egocentric video, in which the visual content tend change continuously because of the mobility of the camera. 
The LSTM baseline consistently achieves second best results with respect to all anticipation times, except for $\tau_a=0.25$, where it is outperformed by EL.
This suggests that the loss functions employed in the RL and EL baselines, originally proposed for early action recognition in third person videos, are not effective in the case of egocentric action anticipation.
{TCN achieves very low performance as compared to most of the considered approach. This suggests that the non-recurrent nature of this approach is not very well suited to the considered anticipation problem, in which it is in general beneficial to refine predictions as more observations are processed.}
The proposed RU model is particularly strong on nouns, achieving a Top-5 noun accuracy of $51.79\%$ and a mean Top-5 noun recall of $49.90\%$, which improves over prior art by $+7.26\%$ and $+7.31\%$ respectively. The small drop in performance between class-agnostic and class-aware measures (i.e., $51.79\%$ vs $49.90\%$) suggests that our method does not over-fit to the distribution of nouns seen during training set. 
It is worth noting that mean Top-5 Recall values are averaged over fairly large sets of $26$ many-shot verbs, $71$ many-shot nouns, and $819$ many-shot actions, as specified in~\cite{Damen2018EPICKITCHENS}. 
Differently, all compared methods obtain large drops in verb and action performance when comparing class-agnostic measures to class-aware measures. 
Our insight into this different pattern is that anticipating the next active object (i.e., anticipating nouns) is much less ambiguous than anticipating the way in which the object will be used (i.e., anticipating verbs and actions). 
It is worth noting that second best Top-5 verb and noun accuracy scores are obtained by different methods (i.e., ATSN in the case of verbs and RL in the case of nouns), while both are outperformed by the proposed RU. 
Despite its low performance when evaluated with class-agnostic measures, ED systematically achieves second best results with respect to mean Top-5 recall and mean $TtA(5)$. This highlights that there is no clear second-best performing method. 
Finally, the mean $TtA(k)$ highlights that the proposed method can anticipate verbs, nouns and actions $1.62$, $1.11$ and $0.76$ seconds in advance respectively.

\tablename~\ref{tab:anticipation_ek_test} compares the performance of the proposed method with baselines and other approaches on the official test sets of the EPIC-Kitchens dataset. 
RU-LSTM outperforms all competitors by consistent margins on both the ``seen'' test, which includes scenes appearing in the training set (\textbf{S1}) and on the ``unseen'' test set, with scenes not appearing in the training set (\textbf{S2}). Also in this case, RU is strong on nouns, obtaining $+6.31\%$ and $+8.23\%$ in \textbf{S1}, as well as $+2.76\%$ and $+2.18$ in \textbf{S2}. Improvements in terms of actions are also significant: $+3.63\%$ and $+5.52\%$ in \textbf{S1}, as well as $+0.92\%$ and $+1.81\%$ on \textbf{S2}.

\begin{table*}
	\caption{Anticipation results on ActivityNet.}
	\label{tab:anticipation_activitynet}
	\begin{adjustbox}{width=0.75\linewidth,center}
		\begin{tabular}{p{1.5cm}ccccccccccc}
			\hline
			\multicolumn{1}{c}{} & \multicolumn{8}{c}{Top-5 Accuracy\% @ different $\tau_a$(s)}   & Top-1\% & M.T-5 Rec.\% & $M.TtA(5)$ \\  \hline %
			\multicolumn{1}{c}{} & $2$ & $1.75$ & $1.5$ & $1.25$ & $1.0$& $0.75$&$0.5$&$0.25$& $1$ & $1$ & / \\ \hline
			DMR~\cite{vondrick2016anticipating} &/ & / & / & / & 52.39 & / & / & / & 24.13 & 39.55 & /\\
			ATSN~\cite{Damen2018EPICKITCHENS}& / & / & / & / & 48.08 & / & / & / & 29.09 & 45.94 & / \\
			TCN~\cite{BaiTCN2018}    & 52.71 & 53.55 & 55.21 & 56.69 & 58.39 & 59.42 & 60.85 & 62.06 & 34.15 & 57.21 & 01.27\\
			ED*~\cite{gao2017red} & 62.89 & 63.71 & 64.23 & 65.27 & 65.84 & 67.09 & 68.16 & 69.79 & 42.83 & 64.98 & 01.40\\
			ED~\cite{gao2017red} & \textbf{70.20} & \textbf{70.45} & \textbf{71.04} & \textbf{71.75} & \textbf{72.93} & \textbf{72.95} & \underline{72.52} & 71.43 & \textbf{48.58} & \textbf{72.32} & \textbf{01.54}\\
			FN~\cite{de2018modeling} & 59.44 & 60.01 & 60.79 & 61.38 & 62.29 & 63.32 & 64.16 & 65.34 & 37.57 & 60.65 & 01.27\\
			RL~\cite{ma2016learning}  & {65.11} & {65.66} & {66.56} & {67.51} & {68.71} & {69.78} & {71.15} & \underline{72.48} & {45.27} & {67.68} & {01.40}\\
			EL~\cite{jain2016recurrent} & 63.21 & 63.99 & 65.15 & 66.03 & 67.05 & 68.25 & 69.49 & 71.15 & 43.06 & 66.17 & 01.37 \\
			LSTM~\cite{lstm2} & 61.61 & 62.92 & 63.72 & 64.31 & 65.70 & 66.71 & 67.87 & 69.24 & 40.37 & 64.44 & 01.34\\
			\textbf{RU-LSTM}  & \underline{65.53} & \underline{66.67} & \underline{67.59} & \underline{69.13} & \underline{70.23} & \underline{71.66} & \textbf{72.73} & \textbf{73.97} & \underline{46.49} & \underline{69.41} & \underline{01.45}\\
			\hline
			Imp & -4.67 & -3.78 & -3.45 & -2.62 & -2.70 & -1.29 & +0.21 & +1.49 & -2.09 & -2.91 & -0.09\\
			\hline
		\end{tabular}
	\end{adjustbox}
\end{table*} 

\subsection{Egocentric Action Anticipation on EGTEA Gaze+}
\label{sec:res_egtea}
\tablename~\ref{tab:anticipation_egtea} reports egocentric action anticipation results on EGTEA Gaze+. The proposed RU approach outperforms the competitors for all anticipation times except for $\tau_a=2s$, in which case its performance is on par with the LSTM baseline. 
Note that the margins of improvement obtained by the proposed method are smaller on EGTEA Gaze+, probably due to its smaller scale ($106$ actions in vs $2,513$ actions in EPIC-KITCHENS). 
Second-best results are achieved by EL for most of the anticipation times, except for $\tau_a\in\{1.25s, 1.0s\}$, where LSTM achieves comparable results. The table shows similar trends to the ones observed in the case of EPIC-Kitchens. 
The methods based on the direct regression of future representations such as DMR and ED still achieve sub-optimal results, especially as compared to other sequence-to-sequence models. 
{Interestingly, ED* achieves better results than ED, which seem to confirm the limited ability of approaches based on direct regression of future representations in the egocentric domain.
Also in this case, the performance of TCN tend to be somewhat limited as compared to recurrent approaches. }
Since no object annotations are available for EGTEA Gaze+, our RU model uses the object detector trained on EPIC-Kitchens for the object branch. Despite the use of object classes not perfectly aligned with the ones contained in the dataset, our approach is strong on nouns even in this case, obtaining an improvement of $+1.80\%$ and $+3.37\%$ with respect to Top-5 accuracy and mean Top-5 recall.

{\subsection{Action Anticipation on ActivityNet}
\label{sec:res_an}
Table~\ref{tab:anticipation_activitynet} reports the results on the third person ActivityNet dataset. 
Interestingly, in this context, ED significantly outperforms ED* and achieves top performances for most of the anticipation times.
This is in line with the findings of the authors of the approach~\cite{gao2017red}, and highlights the different nature of the egocentric scenario as compared to the rather static third person scenario. While the proposed RU model is outperformed by ED on most anticipation times, it systematically achieves second-best performance and has a significant advantage over ED*, which, similarly to the proposed method, does not make use of the unsupervised pre-training.
Differently from previous results, in this case, both RL and EL outperform the LSTM baseline, which suggests that the anticipation losses used in these baseline are more beneficial in the case of third person videos than in the case of first-person videos. This highlights again the differences between the two scenarios. Also in these experiments, TCN achieve worse performance as compared to recurrent models.}

\begin{table}[t]
	\caption{Ablation study on EPIC-KITCHENS.}
	\label{tab:ablation}
	\setlength{\tabcolsep}{5pt}
	\centering
	\begin{adjustbox}{width=\linewidth,center}
		\begin{tabular}{p{2cm}ccccccccc}
			\hline
			\multicolumn{1}{c}{} & \multicolumn{8}{c}{Top-5 ACTION Accuracy\% @ different $\tau_a$(s)} & $TtA$ \\  \hline %
			\multicolumn{1}{c}{} & $2$ & $1.75$ & $1.5$ & $1.25$ & $1.0$& $0.75$&$0.5$&$0.25$ & \\ \hline
			BL (Late)            & \underline{27.96} & \underline{28.76} & \underline{29.99} & \underline{31.09} & \underline{32.02} & \underline{33.09} & \underline{34.13} & \underline{34.92} & \underline{0.66}\\
			RU (Late)            & \textbf{29.10} & \textbf{29.77} & \textbf{31.72} & \textbf{33.09} & \textbf{34.23} & \textbf{35.28} & \textbf{36.10} & \textbf{37.61} & \textbf{0.73} \\ \hline
			Imp.                 & +1.14          & +1.01          & +1.73          & +2.00          & +2.21          & +2.19          & +1.97          & +2.69          & +0.07 \\ \hline
			& 
		\end{tabular}
	\end{adjustbox}
	\centerline{\footnotesize(a) Rolling-Unrolling Mechanism.}
	\vspace{1mm}
	
	\begin{adjustbox}{width=\linewidth,center}
		\begin{tabular}{p{2cm}ccccccccc}
			\hline
			RU (RGB)             & 25.44 & {26.89} & {28.32} & {29.42} & {30.83} & {32.00} & {33.31} & {34.47} & 0.69\\ 
			RU (Flow)            & {17.38} & {18.04} & {18.91} & {19.97} & {21.42} & {22.37} & {23.49} & {24.18} & 0.51 \\ 
			RU (OBJ)             & {24.56} & {25.60} & {26.61} & {28.32} & {29.89} & {30.85} & {31.82} & {33.39} & 0.67 \\ \hline
			
			Early Fusion         & 25.58          & 27.25          & 28.58          & 29.59          & 31.88          & 32.78          & 33.99          & 35.62          & 0.72 \\
			Late Fusion          & \underline{29.10} & \underline{29.77} & \underline{31.72} & \underline{33.09} & \underline{34.23} & \underline{35.28} & \underline{36.10} & \underline{37.61} & \underline{0.73} \\
			MATT                 & \textbf{29.44} & \textbf{30.73} & \textbf{32.24} & \textbf{33.41} & \textbf{35.32} & \textbf{36.34} & \textbf{37.37} & \textbf{38.98} & \textbf{0.76} \\ \hline
			Imp.                 & +0.34          & +0.96          & +0.52          & +0.32          & +1.09          & +1.06          & +1.27          & +1.37          & +0.03\\
			\hline& 
		\end{tabular}
	\end{adjustbox}
	\centerline{\footnotesize (b) Modality Attention Fusion Mechanism.}
	\vspace{1mm}

	\begin{adjustbox}{width=\linewidth,center}
		\begin{tabular}{p{2cm}ccccccccc}
			\hline

			Flow+OBJ & 21.10 & 21.24 & 21.84 & 23.05 & 23.93 & 25.00 & 26.11 & 26.45&0.57\\
			RGB+Flow & 26.75 & 27.43 & 29.20 & 30.15 & 32.16 & 33.49 & 34.37 & 35.46& 0.70\\
			RGB+OBJ & \underline{28.04} & \underline{29.51} & \underline{31.48} & \underline{32.22} & \underline{34.27} & \underline{35.36} & \underline{36.89} & \underline{37.79} & \underline{0.74}\\
			RGB+Flow+OBJ & \textbf{29.44} & \textbf{30.73} & \textbf{32.24} & \textbf{33.41} & \textbf{35.32} & \textbf{36.34} & \textbf{37.37} & \textbf{38.98}& \textbf{0.76}\\

			\hline& 
		\end{tabular}
	\end{adjustbox}
	\centerline{\footnotesize {(c) MATT fusion with different modalities.}}
	\vspace{1mm}
	
	\begin{adjustbox}{width=\linewidth,center}
		\begin{tabular}{p{2cm}ccccccccc}
			\hline
			w/o SCP              & \underline{29.22} & \underline{30.43} & \textbf{32.34} & \underline{33.37} & \underline{34.75} & \underline{35.84} & \underline{36.79} & \underline{37.93} & \underline{0.75} \\
			with SCP             & \textbf{29.44} & \textbf{30.73} & \underline{32.24} & \textbf{33.41} & \textbf{35.32} & \textbf{36.34} & \textbf{37.37} & \textbf{38.98} & \textbf{0.76} \\ \hline
			Imp. of SCP          & +0.22          & +0.30          & -0.10          & +0.04          & +0.57          & +0.50          & +0.58          & +1.05          & +0.01 \\ \hline
		\end{tabular}
	\end{adjustbox}

\vspace{2mm}
	\centerline{\footnotesize (d) Sequence-Completion Pre-training.}
	\vspace{3mm}
	
	\begin{adjustbox}{width=\linewidth,center}
		\begin{tabular}{p{2cm}ccccccccc}
			\hline
			BL (Fusion)          & \underline{27.96} & \underline{28.76} & \underline{29.99} & \underline{31.09} & \underline{32.02} & \underline{33.09} & \underline{34.13} & \underline{34.92} & \underline{0.66} \\
			RU (Fusion)          & \textbf{29.44} & \textbf{30.73} & \textbf{32.24} & \textbf{33.41} & \textbf{35.32} & \textbf{36.34} & \textbf{37.37} & \textbf{38.98} & \textbf{0.76} \\ \hline
			Imp. (Fusion)        & +1.48          & +1.97          & +2.25          & +2.32          & +3.30          & +3.25          & +3.24          & +4.06          & +0.1 \\ \hline
		\end{tabular}
	\end{adjustbox}

\vspace{2mm}
	\centerline{\footnotesize (e) Overall comparison wrt strong baseline.}
\end{table} 

\subsection{Ablation Study on EPIC-Kitchens}
\label{sec:res_ablation}
We performed an ablation study on the EPIC-Kitchens dataset to assess the role of the different components involved in our architecture. Specifically, to assess the role of the proposed rolling-unrolling mechanism, we considered a strong baseline composed of a single LSTM (same configuration as R-LSTM) and three branches (RGB, Flow, OBJ) with late fusion (BL).
Note that, differently from the LSTM baseline compared in the previous sections, this baseline also includes an object branch.
To study the role of rolling-unrolling in isolation, we compare this baseline with respect to a variant of the proposed RU architecture in which MATT has been replaced with late fusion in \tablename~\ref{tab:ablation}(a).
As can be observed, the rolling-unrolling mechanism brings systematic improvements over the strong baseline for all anticipation times.

\begin{figure}
	\includegraphics[width=\linewidth]{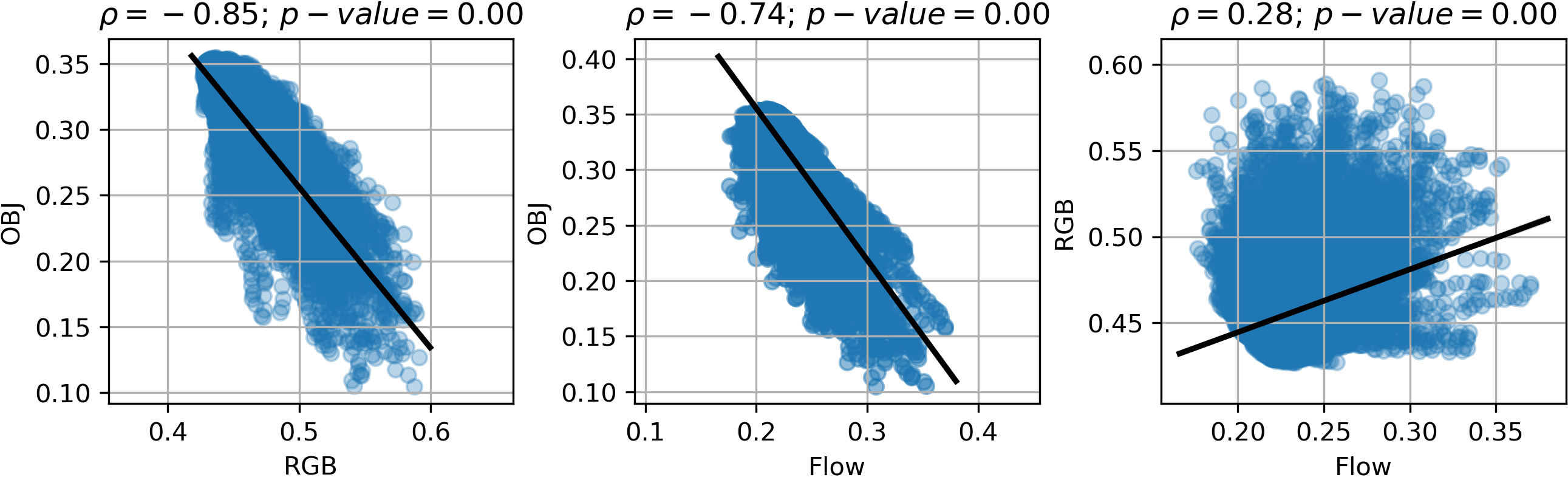}
	\caption{Correlations between modality attention weights}
	\label{fig:correlations}
\end{figure}

\begin{figure}[t]
	\includegraphics[width=\linewidth]{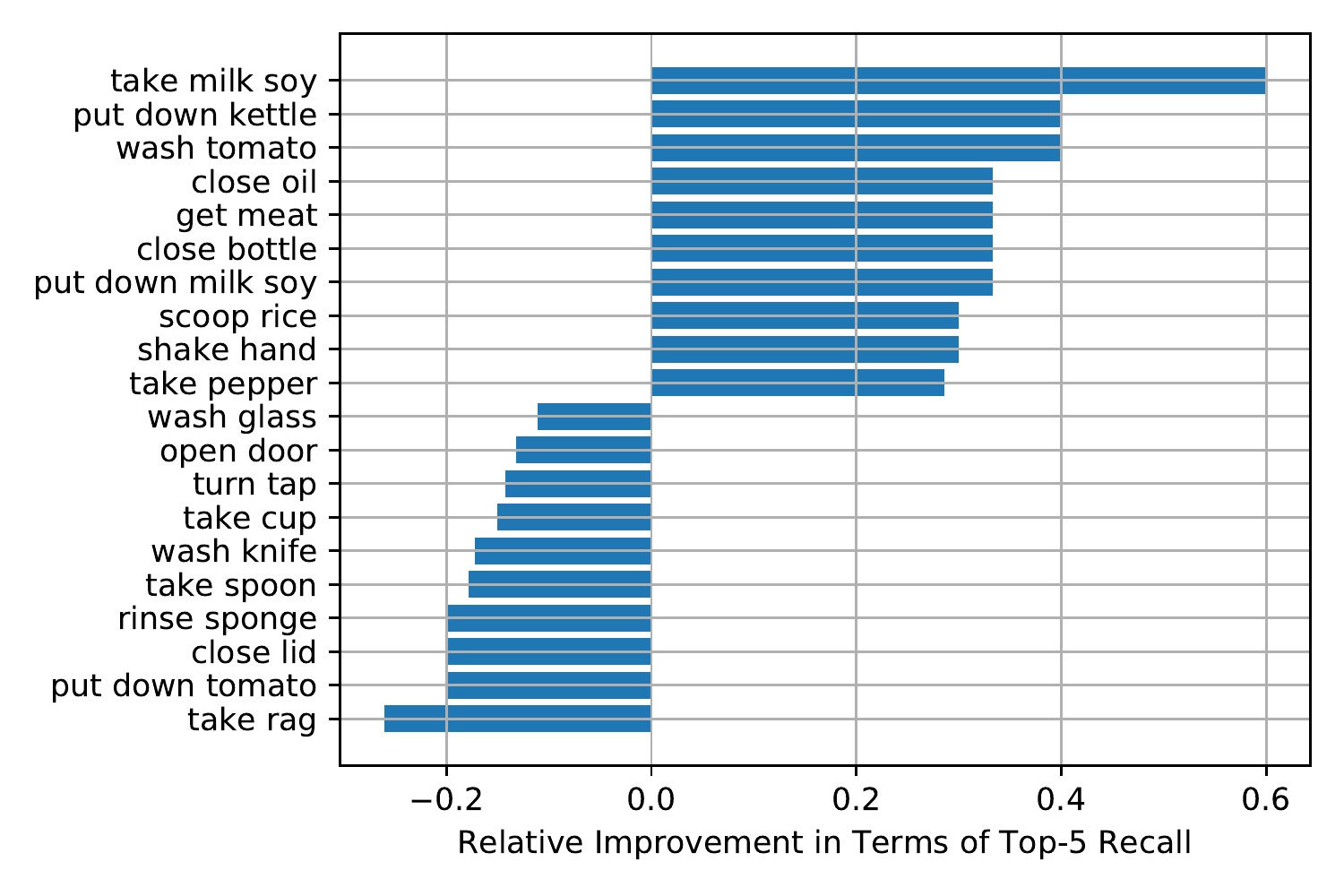}
	\caption{Top-10 and bottom-10 actions which benefited from modality attention in terms of Top-5 Recall.}
	\label{fig:matt_improv}
\end{figure}

In \tablename~\ref{tab:ablation}(b), we study the influence of MATT by comparing it with respect to two standard fusion approaches: early fusion (i.e., feeding the model with the concatenation of the modality-specific representations) and late fusion (i.e., averaging predictions). 
MATT always outperforms late fusion, which consistently achieves second best results, while early fusion always leads to sub-optimal results.
All fusion schemes always improve over the single branches. 
\figurename~\ref{fig:correlations} shows regression plots of the modality attention weights computed by the proposed method on all the samples of the validation set. 
The RGB and OBJ weights are characterized by a strong and steep correlation. 
A similar pattern is observed between Flow and OBJ weights, whereas Flow and RGB weights are characterized by a small positive correlation. 
This suggests that MATT gives more credit to OBJ when RGB and Flow are less informative, whereas it relies on RGB and Flow when the detected objects are not informative. 
{Figure~\ref{fig:matt_improv} shows the top-10 and bottom-10 action categories which benefited from MATT as compared to late fusion, in terms of mean Top-5 Recall. For the analysis, we have considered only classes containing at least $5$ instances in the validation set.
We can observe significant improvements for some actions such as ``take milk soy'' and ``put down kettle'', while there are relatively small negative performance differences with respect to late fusion for some actions such as ``take rag'' and ``put down tomato''.}

{\tablename~\ref{tab:ablation}(c) compares the performances of different versions of the proposed architecture in which MATT is used to fuse different subsets of the considered modalities. Fusing RGB with optical flow (RGB+Flow) or objects (RGB+OBJ) allows to improve over the respective single-branches. Fusing optical flow and objects (Flow+OBJ) improves over the Flow branch, but does not improve over the OBJ branch, while adding RGB (RGB+Flow+OBJ) does allow to improve over the single branches. 
This suggests that the model is not able to take advantage of representations based on optical flow when the RGB signal is not available. 
Interestingly, fusing RGB and objects (RGB+OBJ) allows to obtain better results than fusing RGB and optical flow (RGB+Flow), as it is generally considered in many standard pipelines. This further highlights the importance of objects for egocentric action anticipation. Fusing all modalities leads to the best performance.}

In \tablename~\ref{tab:ablation}(d), we assess the role of Sequence Completion Pre-Training (SCP). The proposed pre-training procedure brings small but consistent improvements for most anticipation times. 
\tablename~\ref{tab:ablation}(e) compares RU with the strong baseline of \tablename~\ref{tab:ablation}(a). 
The comparison shows the cumulative effect of all the proposed procedures/component with respect to a strong baseline which uses three modalities. 
It is worth noting that the proposed architecture brings improvements for all anticipation times, ranging from $+1.48\%$ to $+4.06\%$.

\begin{figure}[t]
	\includegraphics[width=\linewidth]{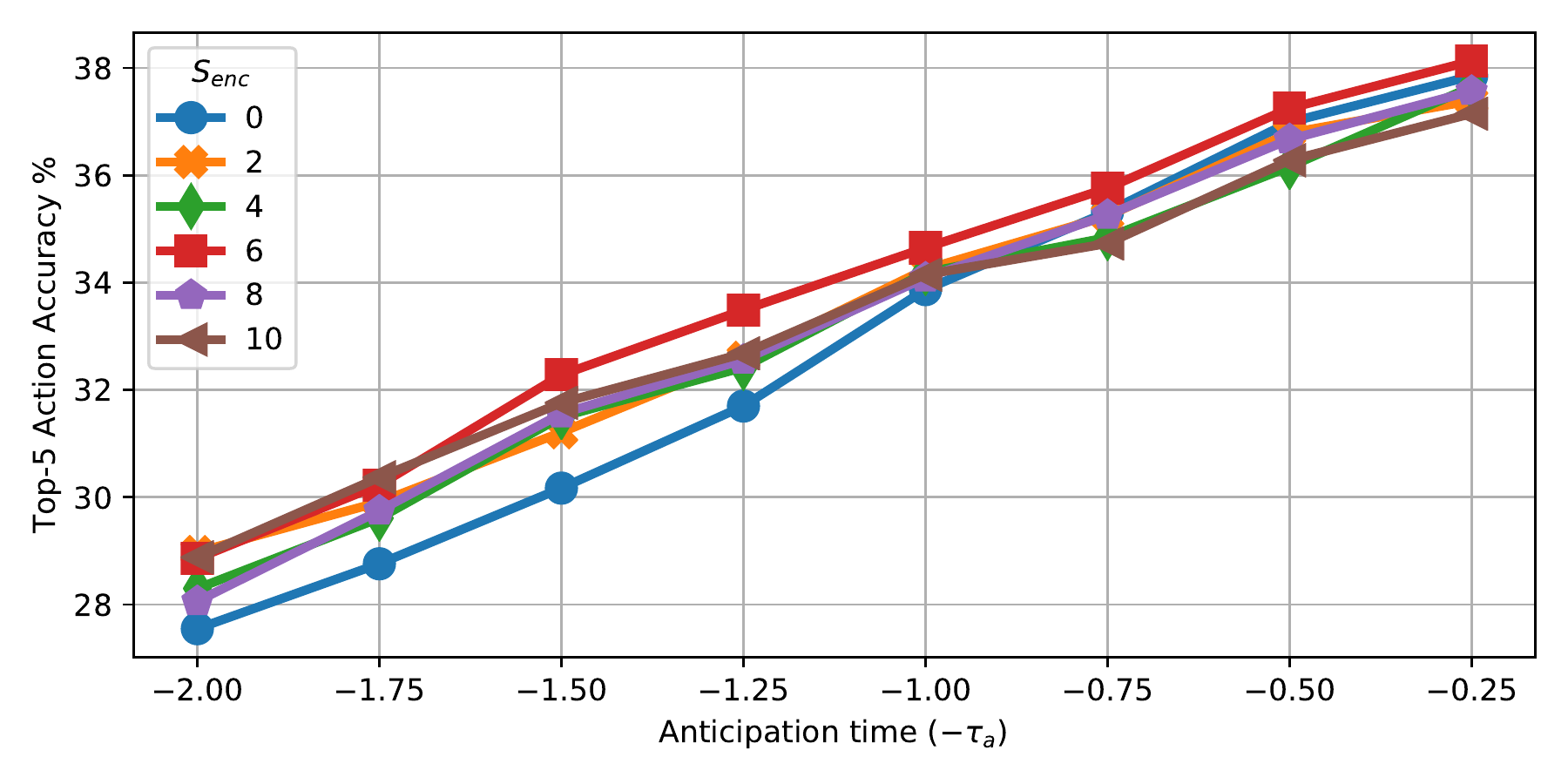}
	\caption{Impact of the choice of $S_{enc}$ on performance.}
	\label{fig:ablation_senc}
\end{figure}

{Figure~\ref{fig:ablation_senc} finally investigates the effect of choosing different values of $S_{enc}$, while the number of anticipation steps is fixed to $S_{ant}=8$. As can be noted, our approach achieves best results across most of the anticipation times for $S_{enc}=6$, while smaller and larger number of encoding steps lead to lower performance. This suggests that, while a sufficiently long temporal context is required to correctly anticipate actions, leveraging very long temporal contexts can be challenging.}

\subsection{Qualitative Results}
\label{sec:res_qualitative}
\begin{figure*}
	\begin{tabular}{c}
		\includegraphics[width=\linewidth,clip=true,trim=0 0 0 90]{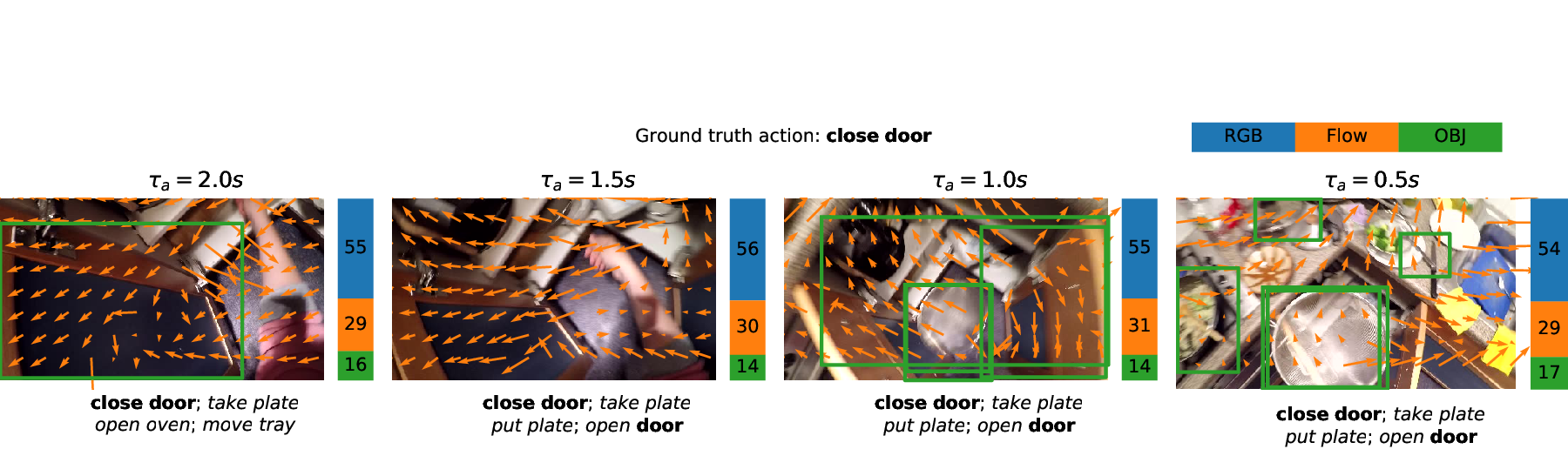}\\
		\includegraphics[width=\linewidth,clip=true,trim=0 0 0 110]{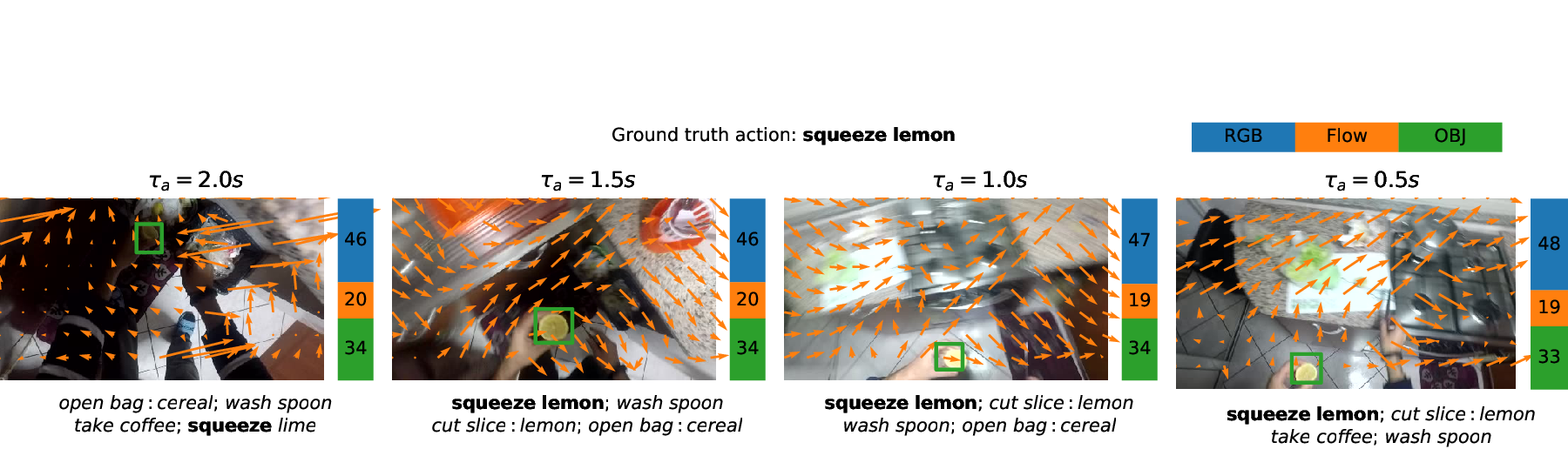}
	\end{tabular}
	\caption{Qualitative examples (best seen on screen). Legend for attention weights: blue - RGB, orange - Flow, green - objects. }
	\label{fig:qualitative}
\end{figure*}

\figurename~\ref{fig:qualitative} reports two qualitative examples of predictions made by the proposed approach at four anticipation times. 
Under each example, are reported the top-4 predictions, whereas modality weights computed by MATT are reported in percentage on the right. 
We show green bounding boxes around the detected objects and orange arrows to illustrate optical flow. 
In the first example (top), the model can predict ``close door'' based on the context and the history of past actions (e.g., taking objects out of the cupboard), hence it assigns large weights to the RGB and Flow modalities and low weights to the OBJ modality. 
In the second example (bottom), the model initially predicts ``squeeze lime'' at $\tau_a=2s$. 
Later, as the lemon is predicted, the prediction is corrected to ``squeeze lemon''. 
Note that in this case the network assigns larger weights to OBJ as compared to the previous example.\footnote{See Appendix~\ref{sec:qualitative} and \textit{https://iplab.dmi.unict.it/rulstm/} for additional examples and videos.}

\section{Additional Results on Early Action Recognition and Action Recognition}
We note that, being a sequence-to-sequence method, our approach can also be employed to perform early action recognition. This is done by processing the video of an action sequentially and outputting a prediction at every time-step. The prediction performed at the last time-step can be also be used action recognition. 

\subsection{Early Action Recognition}
We adapt all sequence-to-sequence models to perform early action recognition by sampling $8$ video snippets form each action segment uniformly and set $S_{enc}=0$, $S_{ant}=8$. 
The models produce predictions at each time-step, corresponding to the observation rates:  $12.5\%$, $25\%$, $37.5\%$, $50\%$, $62.5\%$, $75\%$, $87.5\%$, $100\%$. 
Modality-specific branches are fused by late fusion. 
We compared our approach with respect to the following baselines: FN, RL, EL, {LSTM, TCN}.

\begin{table*}[t]
	\caption{Early recognition results on EPIC-KITCHENS.}
\label{tab:recognition_ek}
	\begin{adjustbox}{width=0.75\linewidth,center}
		\begin{tabular}{p{1.5cm}ccccccccccc}
			\multicolumn{1}{c}{} & \multicolumn{8}{c}{Top-1 ACTION Accuracy\% @ different observation rates} & \multicolumn{3}{c}{$MOR\%$}\\ \hline
			& $12.5\%$ & $25.0\%$ & $37.5\%$ & $50.0\%$ & $62.5\%$& $75.0\%$&$87.5\%$&$100\%$&VERB&NOUN&ACT. \\
			\hline
			{TCN~\cite{BaiTCN2018}}&16.93 & 18.42 & 19.27 & 19.05 & 20.62 & 20.53 & 19.45 & 17.12 & 53.89 & 63.49 & 76.29\\
			FN~\cite{de2018modeling}& 19.61 & 23.85 & 25.66 & 26.85 & 27.47 & 28.34 & 28.26 & 28.38 & 51.12 & 65.12 & 74.49\\
			RL~\cite{ma2016learning} & \underline{22.53} & 25.08 & 27.19 & 28.64 & 29.57 & 30.13 & 30.45 & 30.47 & 51.29 & {63.59} & {73.13}\\
			EL~\cite{jain2016recurrent}& 19.69 & 23.27 & 26.03 & 27.49 & 29.06 & 29.97 & 30.91 & \underline{31.46} & 51.89 & 64.00 & 74.10\\
			{LSTM~\cite{lstm2}}&22.16 & \underline{25.78} & \underline{27.80} & \underline{28.98} & \underline{29.87} & \underline{31.13} & \underline{31.28} & 30.93 & \underline{50.64} & \underline{63.27} & \underline{72.60}\\			 
			\textbf{RU-LSTM} & \textbf{24.48} & \textbf{27.63} & \textbf{29.44} & \textbf{30.93} & \textbf{32.16} & \textbf{33.09} & \textbf{33.63} & \textbf{34.07} & \textbf{47.91} & \textbf{59.15} & \textbf{69.34}\\
			\hline
			Imp.&+1.95 & +1.85 & +1.64 & +1.95 & +2.29 & +1.96 & +2.35 & +2.61 & -2.73 & -4.12 & -3.26\\

			\hline
		\end{tabular}
	\end{adjustbox}
\end{table*} %
\begin{table*}[t]
	\caption{Early recognition results on EGTEA Gaze+.}
\label{tab:recognition_egtea}
	\begin{adjustbox}{width=0.75\linewidth,center}
		\begin{tabular}{p{1.5cm}ccccccccccc}
			\multicolumn{1}{c}{} & \multicolumn{8}{c}{Top-1 ACTION Accuracy\% @ different observation rates} & \multicolumn{3}{c}{$MOR\%$}\\ \hline
			& $12.5\%$ & $25.0\%$ & $37.5\%$ & $50.0\%$ & $62.5\%$& $75.0\%$&$87.5\%$&$100\%$&VERB&NOUN&ACT. \\
			\hline
			{TCN~\cite{BaiTCN2018}} &\underline{49.61} & 52.88 & 55.47 & 56.24 & 56.90 & 57.58 & 57.62 & 56.11 & \underline{34.58} & \underline{33.54} & \underline{45.13}\\
			FN~\cite{de2018modeling}    &44.02 & 50.32 & 53.34 & 55.10 & 56.58 & 57.31 & 57.95 & 57.72 & 37.07 & 37.80 & 48.94\\
			RL~\cite{ma2016learning}   &45.42 & 51.00 & 54.20 & 56.54 & \underline{58.09} & \underline{58.93} & 59.29 & 59.50 & 37.80 & 36.86 & 48.43\\
			EL~\cite{jain2016recurrent} &40.31 & 48.08 & 51.84 & 54.71 & 56.93 & 58.45 & \underline{59.55} & \underline{60.18} & 38.91 & 37.48 & 49.93\\
			{LSTM~\cite{lstm2}} &\textbf{50.22} & \textbf{53.82} & \textbf{55.73} & \textbf{57.20} & 58.01 & 58.79 & 59.09 & 59.32 & 36.20 & 35.46 & {46.89}\\
			\textbf{RU-LSTM} &45.94 & \underline{51.84} & \underline{54.39} & \underline{57.05} & \textbf{58.15} & \textbf{59.31} & \textbf{60.10} & \textbf{60.20} & \textbf{34.91} & \textbf{34.64} & \textbf{45.57}\\
			\hline
			Imp. &-4.28 & -1.98 & -1.34 & -0.15 & 0.06 & 0.38 & 0.55 & 0.02 & 0.33 & 1.10 & 0.44\\
			\hline
		\end{tabular}
	\end{adjustbox}
\end{table*} 

\tablename~\ref{tab:recognition_ek} reports the Top-1 accuracy results obtained by the compared methods with respect to different observation rates on our validation set of EPIC-Kitchens. {$MOR$ scores are reported for verbs, nouns and actions.}
Best results are highlighted in bold numbers. {Note that, differently from Top-1 accuracy, lower $MOR$ scores are better than higher $MOR$ scores, meaning that the method can, in average, recognize an action by observing less frames.}
The proposed method consistently outperforms the competitors at all observation rates by about $+2\%$ in average. 
Interestingly, RU achieves an early action recognition accuracy of $33.09\%$ when observing only $75\%$ of the action, which is already comparable to the accuracy of $34.07\%$ achieved when the full action is observed. 
{Also, the proposed method achieves $MOR$ values lower than the competitors, meaning that, in average, it can predict action correctly by observing less frames.
This suggests that RU can timely recognize actions before they are completed.}
Second-best results are obtained by the LSTM baseline in most cases, indicating that the losses employed by RL and EL are not effective on this dataset for early action recognition.
{Similarly to what observed previously, TCN achieves sub-optimal results as compared to the recurrent approaches.}

\tablename~\ref{tab:recognition_egtea} reports the Top-1 accuracy results on EGTEA Gaze+. The proposed RU is outperformed by the LSTM baseline for observation rates up to $50\%$, while it performs comparably to the competitors for the other observation rates. {Also, the $MOR$ values suggest that the proposed approach predicts actions by observing marginally less video content.} Second-best results are obtained by different methods, and there is not a clear second-best performer in this case. {Interestingly, TCN achieves performances comparable with the recurrent methods on this dataset.}

{\tablename~\ref{tab:recognition_activitynet} reports the early action recognition results obtained by the different methods on ActivityNet. Interestingly, the RL baseline achieves best results or second-best results for most of the observation rates. This suggests that the loss employed by this baseline is effective for early action recognition in the domain of third person videos, which is the scenario for which RL has been originally designed.
As we found the RL loss beneficial, we trained the compared RU method with this loss on this dataset.
The proposed approach achieves performances comparable in average with the other methods, but obtains a $MOR$ smaller by $5.8\%$, which highlights that it can recognize actions by observing $5.8\%$ less video content, in average.
Also in this case, the performance of TCN are lower as compared to the recurrent approaches.}

\begin{table*}[t]
	\caption{Early recognition results on ActivityNet.}
	\label{tab:recognition_activitynet}
	\begin{adjustbox}{width=0.7\linewidth,center}
		\begin{tabular}{p{1.2cm}ccccccccc}
			\multicolumn{1}{c}{} & \multicolumn{8}{c}{Top-1 ACTION Accuracy\% @ different observation rates} & $MOR\%$\\ \hline
			& $12.5\%$ & $25.0\%$ & $37.5\%$ & $50.0\%$ & $62.5\%$& $75.0\%$&$87.5\%$&$100\%$& \\
			\hline
			TCN~\cite{BaiTCN2018} &50.03 & 54.85 & 56.73 & 58.92 & 60.60 & 61.55 & 61.51 & 60.71 & {41.65}\\
			FN~\cite{de2018modeling} &53.74 & 58.94 & 61.82 & 63.82 & 65.51 & 66.22 & 67.00 & 67.68 & 42.69\\
			RL~\cite{ma2016learning} &55.44 & \textbf{62.41} & \textbf{65.23} & \textbf{67.40} & \textbf{68.65} & \underline{69.83} & \underline{70.35} & \underline{70.56} & 39.23\\
			EL~\cite{jain2016recurrent} &55.51 & 61.74 & 64.92 & 66.69 & 68.22 & 68.98 & 69.66 & 70.50 & 39.22\\
			LSTM~\cite{lstm2} &\textbf{55.99} & {61.91} & \underline{65.05} & \underline{67.00} & {68.31} & 69.07 & 69.45 & 70.20 & \underline{39.16}\\
			\textbf{RU+RL}&\underline{55.74} & \underline{62.01} & 64.81 & 66.67 & \underline{68.58} & \textbf{70.27} & \textbf{70.69} & \textbf{71.17} & \textbf{36.89}\\
			\hline
			Imp.&-0.25 & -0.40 & -0.42 & -0.73 & -0.07 & +0.44 & +0.34 & +0.61 & -5.80\\

			\hline
		\end{tabular}
	\end{adjustbox}
\end{table*}

\begin{table*}[t]
		\caption{Egocentric action recognition results on the EPIC-Kitchens test set.}
	\label{tab:recognition_ek_test}
	\begin{adjustbox}{width=\linewidth,center}
		\setlength{\tabcolsep}{3pt}
		\begin{tabular}{llccc|ccc|ccc|ccc}
			& & \multicolumn{3}{c|}{Top-1 Accuracy\%} & \multicolumn{3}{c|}{Top-5 Accuracy\%} & \multicolumn{3}{c|}{Avg Class Precision\%} & \multicolumn{3}{c}{Avg Class Recall\%} \\ \hline
			& & VERB & NOUN & ACTION & VERB & NOUN & ACTION & VERB & NOUN & ACTION & VERB & NOUN & ACTION \\ \hline
			\multirow{13}{*}{\rotatebox{90}{\textbf{S1}}} &
			2SCNN~\cite{Damen2018EPICKITCHENS}& 42.16 & 29.14 & 13.23 & 80.58 & 53.70 & 30.36 & 29.39 & 30.73 & 05.53 & 14.83 & 21.10 & 04.46\\
			&TSN~\cite{Damen2018EPICKITCHENS} & 48.23 & 36.71 & 20.54 & 84.09 & 62.32 & 39.79 & 47.26 & 35.42 & 10.46 & 22.33 & 30.53 & 08.83\\
			&LSTA~\cite{sudhakaran2018lsta} & 59.55 & 38.35 & 30.33 & 85.77 & 61.49 & 49.97 & 42.72 & 36.19 & 14.46 & 38.12 & 36.19 & 17.76\\
			&VNMCE~\cite{furnari2018Leveraging} & 54.22 & 38.85 & 29.00 & 85.22 & 61.80 & 49.62 & 53.87 & 38.18 & 18.22 & 35.88 & 32.27 & 16.56\\
			&{TRN~\cite{zhou2018temporal}} & 61.12 & 39.28 & 27.86 & 87.71 & 64.36 & 47.56 & 52.32 & 35.68 & 16.38 & 32.93 & 34.18 & 14.36\\
			&{TSM~\cite{lin2019tsm}} & 62.37 & 41.88 & 29.90 & \underline{88.55} & 66.43 & 49.81 & \textbf{59.51} & 39.50 & 18.38 & 34.44 & 36.04 & 15.80\\
			&{TBN~\cite{kazakos2019epic}} & \textbf{64.75} & \underline{46.03} & \underline{34.80} & \textbf{90.70} & \underline{71.34} & \textbf{56.65} & \underline{55.67} & \textbf{43.65} & \textbf{22.07} & \textbf{45.55} & 42.30 & \textbf{21.31}\\
			&{Miech et al.~\cite{miech2019leveraging}} & 43.51 & 32.94 & 20.19 & 84.38 & 61.66 & 43.57 & 28.42 & 27.99 & 07.62 & 24.18 & 26.83 & 08.85\\
			&{FAIR~\cite{ghadiyaram2019large}} & \underline{64.14} & \textbf{47.65} & \textbf{35.75} & 87.64 & 70.66 & 54.65 & 43.64 & \underline{40.53} & \underline{18.95} & \underline{38.31} & \textbf{45.29} & \underline{21.13}\\
			&{FB~\cite{wu2019long}} & 60.00 & 45.00 & 32.70 & 88.40 & \textbf{71.80} & \underline{55.32} &  / & / & / & / & / & /\\
			
			&RU-LSTM & 56.93 & 43.05 & 33.06 & 85.68 & 67.12 & \underline{55.32} & 50.42 & 39.84 & 18.91 & 37.82 & \underline{38.11} & 19.12\\
			 \hline
			&Imp. & -07.82 & -04.60 & -02.69 & -05.02 & -04.68 & -01.33 & -09.09 & -03.81 & -03.16 & -07.73 & -07.18 & -02.19\\
			
			\hline
			\multirow{13}{*}{\rotatebox{90}{\textbf{S2}}} 
			&2SCNN~\cite{Damen2018EPICKITCHENS} & 36.16 & 18.03 & 07.31 & 71.97 & 38.41 & 19.49 & 18.11 & 15.31 & 02.86 & 10.52 & 12.55 & 02.69\\
			&TSN~\cite{Damen2018EPICKITCHENS} & 39.40 & 22.70 & 10.89 & 74.29 & 45.72 & 25.26 & 22.54 & 15.33 & 05.60 & 13.06 & 17.52 & 05.81\\
			&LSTA~\cite{sudhakaran2018lsta} & 47.32 & 22.16 & 16.63 & 77.02 & 43.15 & 30.93 & \underline{31.57} & 17.91 & 08.97 & \underline{26.17} & 17.80 & 11.92\\
			&VNMCE~\cite{furnari2018Leveraging} & 40.90 & 23.46 & 16.39 & 72.11 & 43.05 & 31.34 & 26.62 & 16.83 & 07.10 & 15.56 & 17.70 & 10.17\\
			&{TRN~\cite{zhou2018temporal}} & 51.62 & 26.02 & 17.34 & 78.42 & 48.99 & 32.57 & \textbf{32.47} & 19.99 & 09.45 & 21.63 & 21.53 & 10.11\\
			&{TSM~\cite{lin2019tsm}} & 51.96 & 25.61 & 17.38 & 79.21 & 49.47 & 32.67 & 27.43 & 17.63 & 09.17 & 20.19 & 22.93 & 11.18\\
			&{TBN~\cite{kazakos2019epic}} & \underline{52.69} & 27.86 & 19.06 & \underline{79.93} & 53.78 & 36.54 & 31.44 & \underline{21.48} & 12.00 & \textbf{28.21} & \underline{23.53} & 12.69\\
			&{Miech et al.~\cite{miech2019leveraging}} & 39.30 & 22.43 & 14.10 & 76.41 & 47.35 & 32.43 & 20.42 & 15.96 & 04.83 & 16.95 & 17.72 & 08.46\\
			&{FAIR~\cite{ghadiyaram2019large}} & \textbf{55.24} & \textbf{33.87} & \textbf{23.93} & \textbf{80.23} & \textbf{58.25} & \textbf{40.15} & 25.71 & \textbf{28.19} & \textbf{15.72} & 25.69 & \textbf{29.51} & \textbf{17.06}\\
			&{FB~\cite{wu2019long}} & 50.90 & \underline{31.50} & 21.20 & 77.60 & \underline{57.80} & \underline{39.40} & / & / & / & / & / & /\\
			&RU-LSTM & 43.67 & 26.77 & \underline{19.49} & 73.30 & 48.28 & 37.15 & 23.40 & 20.82 & \underline{09.72} & 18.41 & 21.59 & \underline{13.33}\\ \hline
			&Imp. & -11.57 & -07.10 & -04.44 & -06.93 & -09.97 & -03.00 & -09.07 & -07.37 & -06.00 & -09.80 & -07.92 & -03.73\\
			\hline

		\end{tabular}
	\end{adjustbox}

\end{table*}

\subsection{Action Recognition}
We finally compare the performance of our method with respect to other state-of-the-art approaches on the task of action recognition on EPIC-Kitchens and EGTEA Gaze+. 
We do not assess the performance of our approach on ActivityNet as this dataset is generally used by the community for action localization rather than recognition.
Although our method does not generally outperform the competitors, it achieves competitive results in some cases.

\tablename~\ref{tab:recognition_ek_test} compares the performance of the proposed method with the state-of-the-art approaches to egocentric action recognition on the two test sets of EPIC-Kitchens. 
Being designed for early egocentric action anticipation, the proposed RU approach does not outperform the competitors, but achieves competitive results with the state-of-the-art, obtaining $-4.44\%$ and $-3\%$ on Top-1 and Top-5 action accuracy. Also, it outperforms several action recognition baselines such as TSN, 2SCNN, TRN and TSM on Top-1 and Top-5 action accuracy. 

Table~\ref{tab:recognition_egtea_test} reports the action recognition results on EGTEA Gaze+. Despite being designed for action anticipation, RU outperforms recent approaches, such as Li et al.~\cite{Li_2018_ECCV} ($+6.9\%$ wrt $53.3\%$) and Zhang et al.~\cite{zhang2018adding} ($+3.19\%$ wrt $57.01\%$ - reported from~\cite{sudhakaran2018lsta}), and obtains performances comparable to state-of-the-art approaches such as Sudhakaran and Lanz~\cite{sudhakaran2018attention} ($-0.56\%$ wrt $60.76$) and Sudhakaran et al.~\cite{sudhakaran2018lsta} ($-1.66\%$ wrt $61.86\%$).

\begin{table}[t]
		\caption{Recognition results on EGTEA Gaze+.}
	\label{tab:recognition_egtea_test}
	\begin{adjustbox}{width=0.65\linewidth,center}
		\begin{tabular}{l|c|c}
			\hline
			Method & Acc.\% & Imp. \\ \hline
			Lit et al.~\cite{li2015delving} &	46.50 &	+13.7 \\
			Li et al.~\cite{Li_2018_ECCV} & 53.30	& +6.90\\
			Two stream~\cite{simonyan2014two}	& 41.84 &	+18.7\\
			I3D~\cite{carreira2017quo}	& 51.68 &	+8.52\\
			TSN~\cite{wang2016temporal}	& 55.93 &	+4.27\\
			eleGAtt~\cite{zhang2018adding}	& 57.01&	+3.19\\
			ego-rnn~\cite{sudhakaran2018attention}	& \underline{60.76}&	-0.56\\
			LSTA~\cite{sudhakaran2018lsta}	& \textbf{61.86}&	-1.66\\
			\textbf{RU-LSTM}	& 60.20 & /	\\
			\hline
		\end{tabular}
	\end{adjustbox}
\end{table}

\section{Conclusion}
\label{sec:conclusion}
We presented Rolling-Unrolling LSTMs, an architecture for egocentric action anticipation. The proposed architecture includes two separate LSTMs, designed to explicitly disentangle two sub-tasks: summarizing the past (encoding) and predicting the future (inference). To encourage such disentanglement, the architecture is trained with a novel sequence-completion pre-training. A modality attention network is introduced to fuse multi-modal predictions obtained by three branches processing RGB frames, optical flow fields and object-based features. Experiments on three benchmark datasets highlight that the proposed approach achieves state-of-the-art results on the task of action anticipation on both first-person and third-person scenarios, and generalizes to the tasks of early action recognition and action recognition. To encourage research on the topic, we publicly released the source code of the proposed approach, together with pre-trained models and extracted features at our project web page: \textit{http://iplab.dmi.unict.it/rulstm}. 
\ifCLASSOPTIONcompsoc
  \section*{Acknowledgments}
\else
  \section*{Acknowledgment}
\fi
This research is supported by Piano della Ricerca 2016-2018, linea di
Intervento 2 of DMI, University of Catania and MIUR AIM - Attrazione e Mobilità Internazionale Linea 1 - AIM1893589 - CUP E64118002540007

\ifCLASSOPTIONcaptionsoff
  page
\fi

\setcounter{section}{0}
\section*{Appendix}
\renewcommand{\thesection}{\Alph{section}}
\renewcommand{\thesubsection}{\Alph{subsection}}

\section{Implementation and Training Details of the Proposed Method}
\label{sec:implementation_details}
This section reports implementation and training details of the different components involved in the proposed method. The reader is also referred to the code available online for the implementation of the proposed approach: \textit{https://iplab.dmi.unict.it/rulstm/}.

\subsection{Architectural Details}
We use a Batch Normalized Inception architecture~\cite{ioffe2015batch} in the representation functions $\phi_1$ and $\phi_2$ of the spatial and motion branches.
For the object branch, we use a Faster R-CNN object detector~\cite{girshick2015fast} with a ResNet-101 backbone~\cite{he2016deep}, as implemented in~\cite{Detectron2018}. 
Both the Rolling LSTM (R-LSTM) and the Unrolling LSTM (U-LSTM) contain a single hidden layer with $1024$ units. Dropout with $p=0.8$ is applied to the input of each LSTM and to the input of the final fully connected layer used to obtain class scores. 
The Modality ATTention network (MATT) is a feed-forward network with three fully connected layers containing respectively $h/4$, $h/8$ and $3$ hidden units, where $h$ is the dimension of the input to the attention network (i.e., the concatenation of the hidden and cell states of $1024$ units of all modality-specific R-LSTMs). 
When three modalities are considered, we obtain an input of dimension  $h=3 \cdot 2 \cdot 1024 = 6144$.
Dropout with $p=0.8$ is applied to the input of the second and third layers of the attention network to avoid over-fitting. ReLU activations are used within the attention network.

\subsection{Training Procedure}
We train the spatial and motion CNNs for the task of egocentric action recognition with TSN~\cite{wang2016temporal}.
We set the number of segments of TSN to $3$ and train the model with Stochastic Gradient Descent (SGD) and standard cross entropy for $160$ epochs with an initial learning rate equal to $0.001$, which is decreased by a factor of $10$ after $80$ epochs. 
We use a mini-batch size of $64$ samples and train the models on a single Titan X GPU.
For all other parameters, we use the values recommended in~\cite{wang2016temporal}. 
We train the object detector to recognize the $352$ object classes of the EPIC-Kitchens dataset. 
We use the same object detector trained on EPIC-Kitchens when performing experiments on EGTEA Gaze+, as the latter dataset does not contain object bounding box annotations. 
We do not use an object branch in the case of ActivityNet.
This training procedure allows to learn the parameters $\theta^1$, $\theta^2$ and $\theta^3$ of the representation functions related to the three modalities (i.e., RGB, Flow, OBJ). 
After this procedure, these parameters are fixed and they are no further optimized. For efficiency, we pre-compute representations over the whole dataset.

Each branch of the RU-LSTM is trained with SGD and the cross entropy loss with a fixed learning rate equal to $0.01$ and a momentum equal to $0.9$.
The loss is averaged both across the samples of the mini-batches and across the predictions obtained at different time-stamps. 
Each branch is first pre-trained with Sequence Completion Pre-training (SCP). 
Specifically, appearance and motion branches are trained for $100$ epochs, whereas the object branch is trained for $200$ epochs. 
The branches are then fine-tuned for the action anticipation task. 
Once each branch has been trained, the complete architecture with three branches is assembled using MATT to form a three-branch network and the model is further fine-tuned for $100$ epochs using cross entropy and the same learning parameters. 
In the case of early action recognition, each branch is trained for $200$ epochs (both SCP and main task) with a fixed learning rate equal to $0.01$ and momentum equal to $0.9$.

We apply early stopping at each training stage. This is done by choosing the iterations of the intermediate and final models which obtain the best Top-5 action anticipation accuracy for the anticipation time $\tau_a=1s$ on the validation set. 
In the case of early action recognition, we choose the epoch in which we obtain the best average Top-1 action accuracy across observation rates. 
The same early stopping strategy is applied to all the methods for fair comparison. 
The proposed architecture has been implemented using the PyTorch library~\cite{paszke2019pytorch}.

\section{Implementation and Training Details of the Compared Methods}
\label{sec:implementation_details_compared}
Since no official public implementations are available for the compared methods, we performed experiments using our own implementations. In this section, we report the implementation details of each of the compared method.

\subsection{Deep Multimodal Regressor (DMR)}
We implemented the Deep Multimodal Regressor proposed in~\cite{vondrick2016anticipating} setting the number of multi-modal branches with interleaved units to $k=3$. 
For fair comparisons, we substituted the AlexNet backbone originally considered in~\cite{vondrick2016anticipating} with a BNInception CNN pre-trained on ImageNet. 
The CNN has been trained to anticipate future representations extracted using BNInception pre-trained on ImageNet using the procedure proposed by the authors. 
Specifically, we performed mode update every epoch. Since training an SVM with large number of classes is challenging (we have $2,513$ different action classes in the case of EPIC-Kitchens), we substituted the SVM with a Multi Layer Perceptron (MLP) with $1024$ hidden units and dropout with $p=0.8$ applied to the input of the first and second layer. 
To comply with the pipeline proposed in~\cite{vondrick2016anticipating}, we pre-trained the model in an unsupervised fashion and then trained the MLP separately on representations pre-extracted from the training set using the optimal modes found at training time. 
As a result, during the training of the MLP, the weights of the CNN are not optimized. 
The DMR architecture has been trained with Stochastic Gradient Descent using a fixed learning rate equal to $0.1$ and a momentum equal to $0.9$. 
The network has been trained for several epochs until the validation loss saturated. Note that training the CNN on the EPIC-Kitchens dataset takes several days on a single Titan X GPU using our implementation. 
After training the DMR, we applied early stopping by selecting the iteration with the lowest validation loss. 
The MLP has then been trained with Stochastic Gradient Descent with fixed learning rate equal to $0.01$ and momentum equal to $0.9$. Early stopping has been applied also in this case considering the iteration of the MLP achieving the highest Top-5 action accuracy on the validation set.

\subsection{Anticipation TSN (ATSN)}
We implemented this model following~\cite{Damen2018EPICKITCHENS}. Specifically, the model has been trained using TSN with 3 segments. We modified the network to output verb and noun scores and trained it summing the cross entropy losses applied independently to verbs and nouns. 
At test time, we obtained action probabilities by assuming conditional independence of verbs and nouns given the sample as follows: $p(a=(v,n)|x)=p(v|x)\cdot p(n|x)$, where $a=(v,n)$ is an action involving verb $v$ and noun $n$, $x$ is the input sample, whereas $p(v|x)$ and $p(n|x)$ are the verb and noun probabilities computed directly by the network.

\subsection{ATSN + VNMCE Loss (MCE)}
This method has been implemented training the TSN architecture used in the case of ATSN with the Verb-Noun Marginal Cross Entropy Loss proposed in~\cite{furnari2018Leveraging}. We used the official code available at \textit{https://github.com/fpv-iplab/action-anticipation-losses/}.

\subsection{Encoder-Decoder LSTM (ED)}
We implemented this model following the details specified in~\cite{gao2017red}. 
For fair comparison with respect to the proposed method, the model takes RGB and Flow features obtained using the representation functions considered for the RGB and Flow branches of our RU architecture. 
Differently from~\cite{gao2017red}, we do not include a reinforcement learning term in the loss as our aim is not to distinguish the action from the background as early as possible as proposed in~\cite{gao2017red}. 
The hidden state of the LSTMs is set to $2048$ units. The model encodes representations for $16$ steps, while decoding is carried out for $8$ steps at a step-size of $0.25s$. 
The architecture is trained in two stages. In the first stage, the model is trained to predict future representations. This stage is carried out for $100$ epochs. The training data for a given epoch is obtained by sampling $100$ random sequences for each video in the case of EPIC-Kitchens and EGTEA Gaze+ and $10$ random sequences for each video in the case of ActivityNet. The difference in the number of sampled sequences reflects the different natures of the datasets: EPIC-Kitchens and EGTEA Gaze+ tend to contain few long videos, while ActivityNet tends to contain many shorter videos. The sequences are re-sampled at each epoch, which reduces the risk of overfitting. In all cases, the models converged within $100$ epochs. Similarly to the other approaches, we apply early stopping by selecting the model with the lowest validation loss.
In the second stage, the architecture is fine-tuned to predict future representations and anticipate future actions for $100$ epochs.
In both stages we use the Adam optimizer and a learning rate of $0.001$ as suggested by the authors of~\cite{gao2017red}.
We also compare this method with a version of the approach in which the unsupervised pre-training stage is skipped (termed ED*). In this case, only the second stage of training is performed.

\subsection{Feedback-Network LSTM (FN)}
The method proposed in~\cite{de2018modeling} has been implemented considering the best performing architecture among the ones investigated by the authors. This architecture comprises the ``optional'' LSTM layer and performs fusion by concatenation. 
The network uses our proposed video processing strategy. For fair comparison, we implemented the network as a two-stream architecture with two branches processing independently RGB and Flow features. 
The final predictions have been obtained with late fusion (equal weights for the two modalities). 
For fair comparisons, we used the representation functions of our architecture to obtain RGB and Flow features. 
The model has hidden layers of $1024$ units, which in our experiments leaded to improved results with respect to the $128$ features proposed by the authors~\cite{de2018modeling}. 
The model has been trained using the same parameters used in the proposed architecture.

\subsection{Temporal Convolutional Network (TCN)}
This baseline has been implemented using the code provided by the authors~\cite{BaiTCN2018}. Similarly to FN, this method uses the our proposed video processing strategy. The network has $5$ layers with kernels of size $7$ in each layer. A dropout with $p=0.8$ is applied to the input of each layer of the network. The model has been trained using cross entropy.

\subsection{LSTM, RL \& EL}
These three methods have been implemented considering a single LSTM with the same parameters of our Rolling LSTM. Similarly to FN, the models have been trained as two-stream models with late fusion used to obtain final predictions (equal weights). 
The input RGB and Flow features have been computed using the representation functions considered in our architecture. 
The models have been trained with the same parameters used in the proposed architecture. 
LSTM has been trained using cross entropy, RL has been trained using the ranking loss on the detection score proposed in~\cite{ma2016learning}, whereas EL ihas been trained using the exponential anticipation loss proposed in~\cite{jain2016recurrent}.

\begin{figure*}
	\includegraphics[width=\linewidth,clip=true,trim=0 64 0 60]{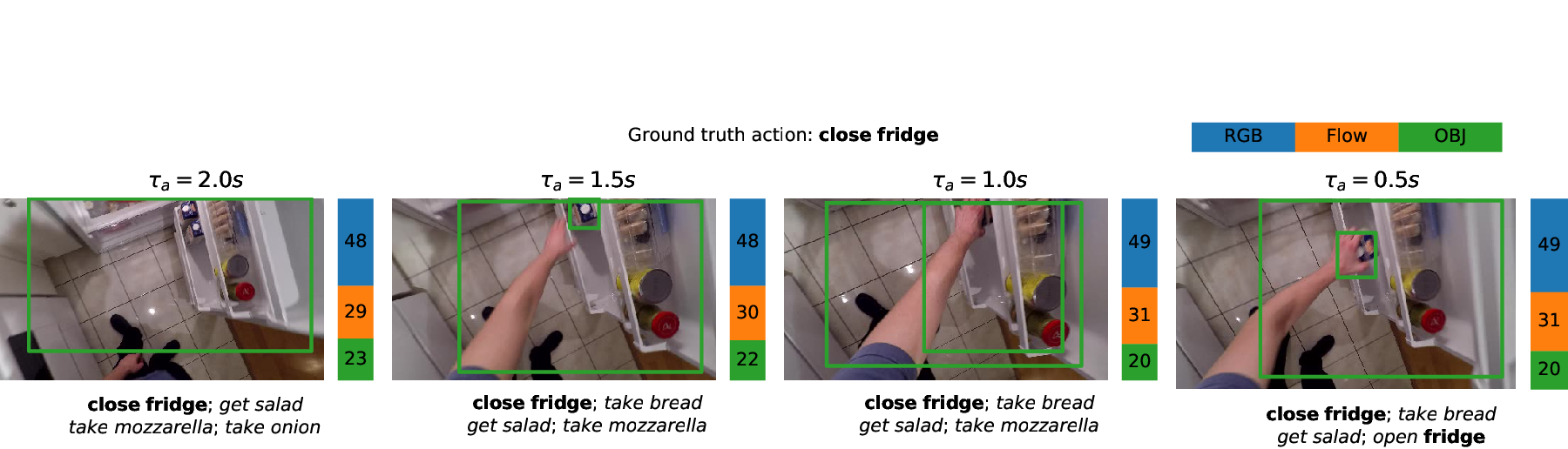}\\
	\includegraphics[width=\linewidth,clip=true,trim=0 0 0 190]{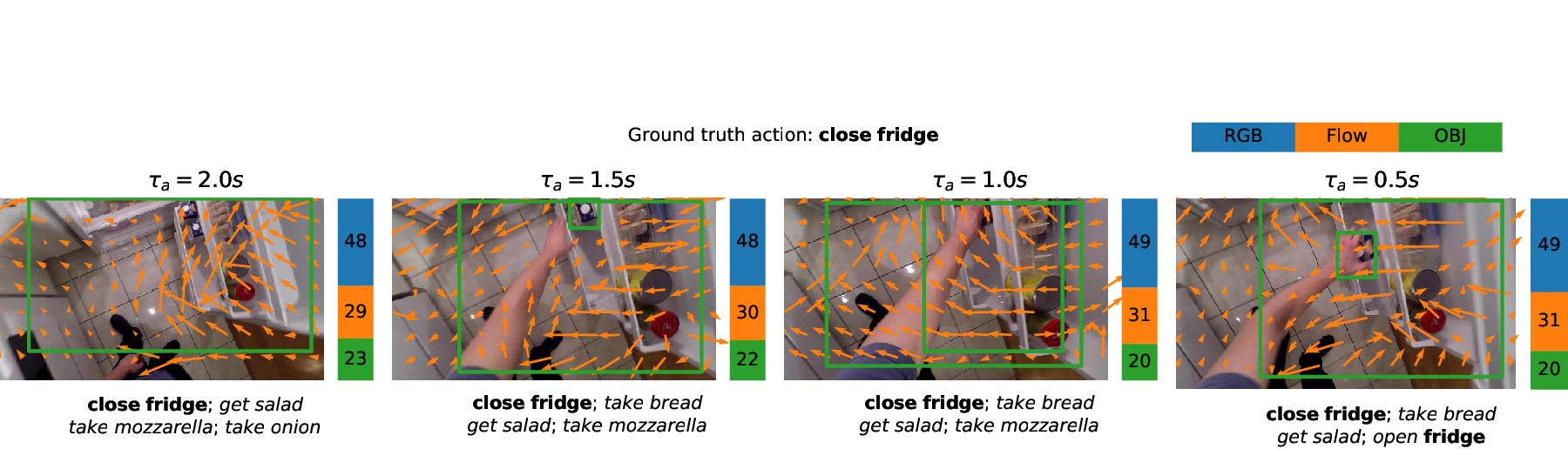}
	
	\includegraphics[width=\linewidth,clip=true,trim=0 64 0 60]{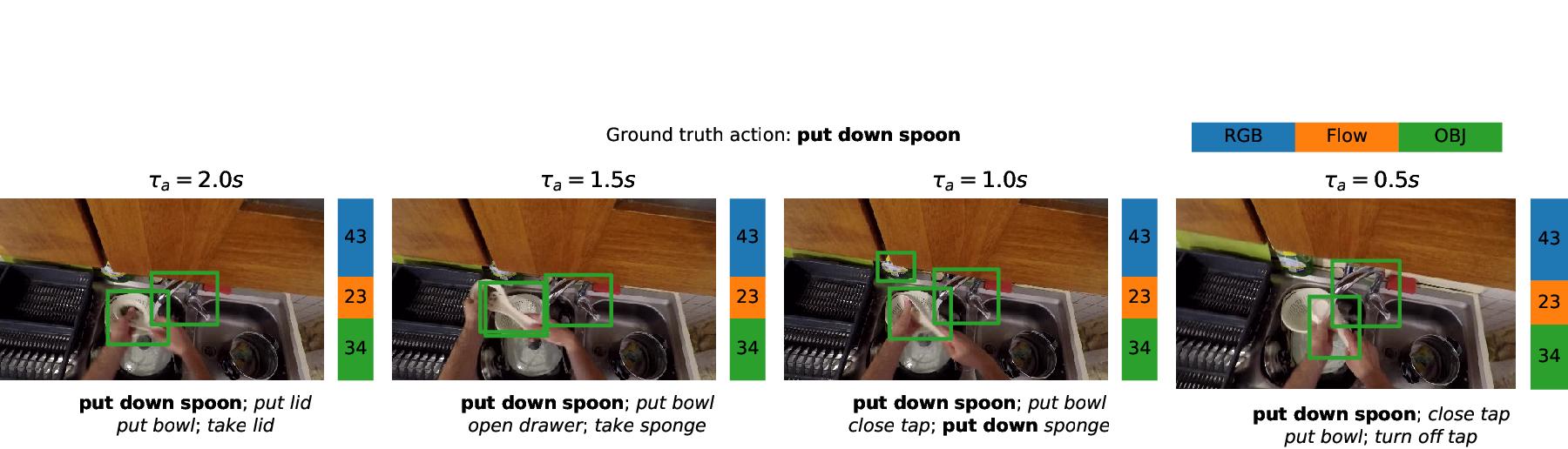}\\
	\includegraphics[width=\linewidth,clip=true,trim=0 0 0 190]{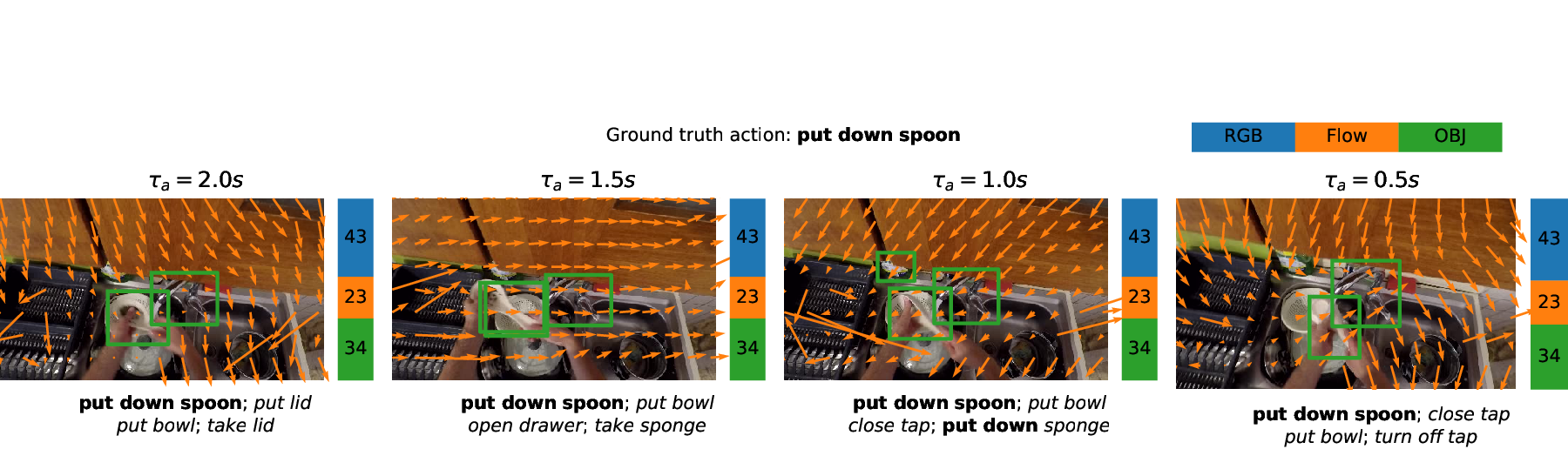}
	
	\includegraphics[width=\linewidth,clip=true,trim=0 64 0 60]{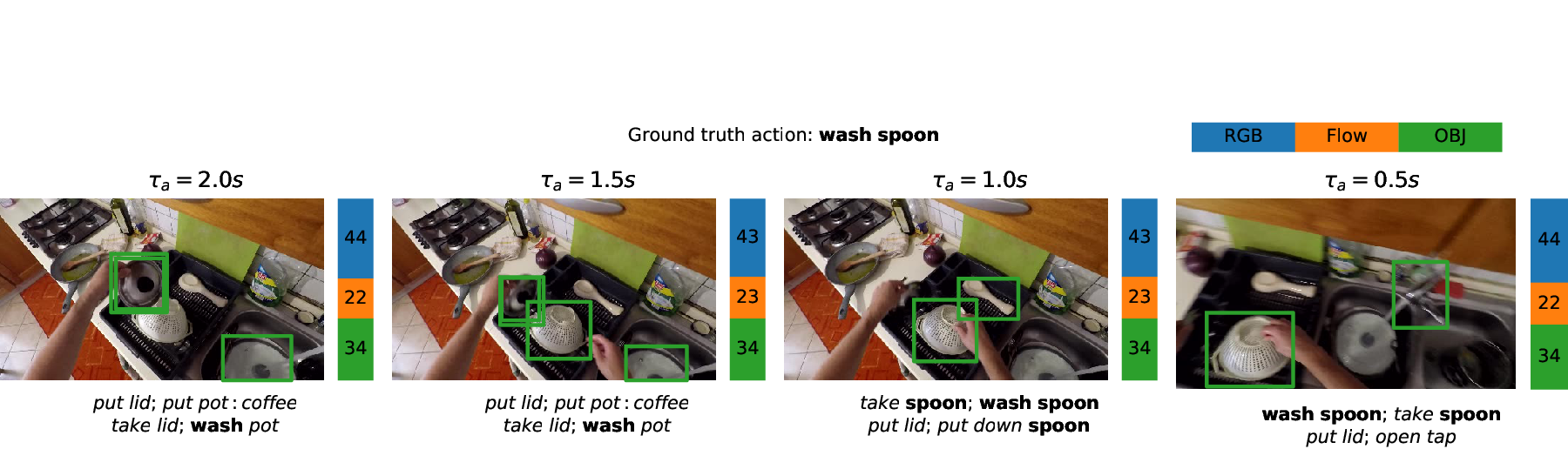}\\
	\includegraphics[width=\linewidth,clip=true,trim=0 0 0 190]{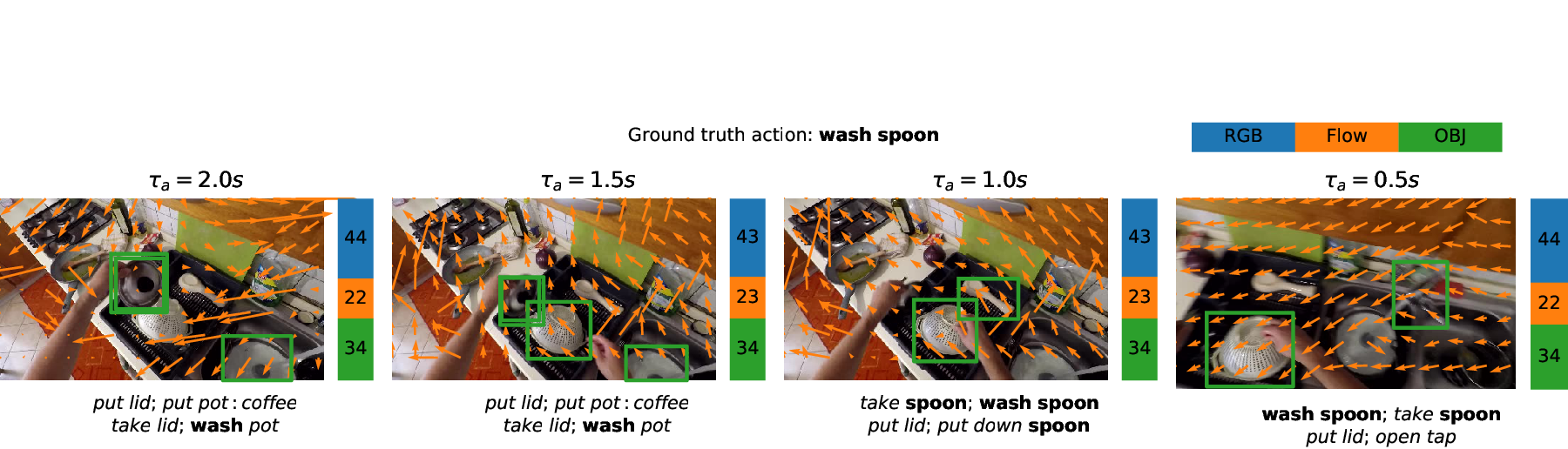}

	\caption{Success action anticipation example qualitative results (best seen on screen).}
	\label{fig:qualitative_success}
\end{figure*}

\begin{figure*}
	\includegraphics[width=\linewidth,clip=true,trim=0 64 0 60]{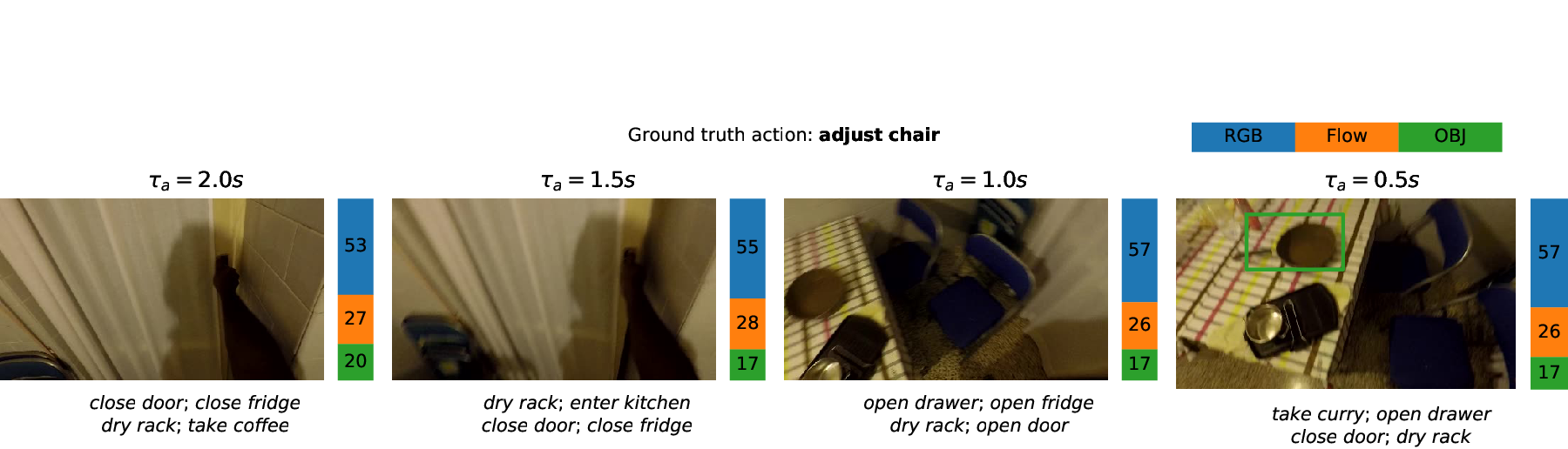}\\
	\includegraphics[width=\linewidth,clip=true,trim=0 0 0 190]{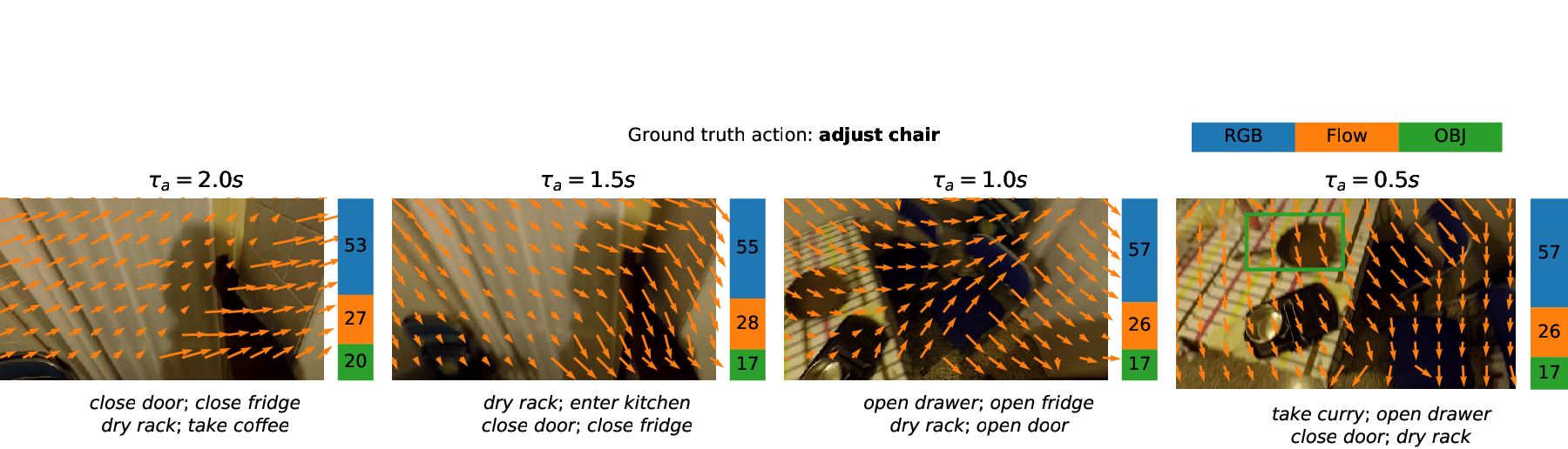}
	
	\includegraphics[width=\linewidth,clip=true,trim=0 64 0 60]{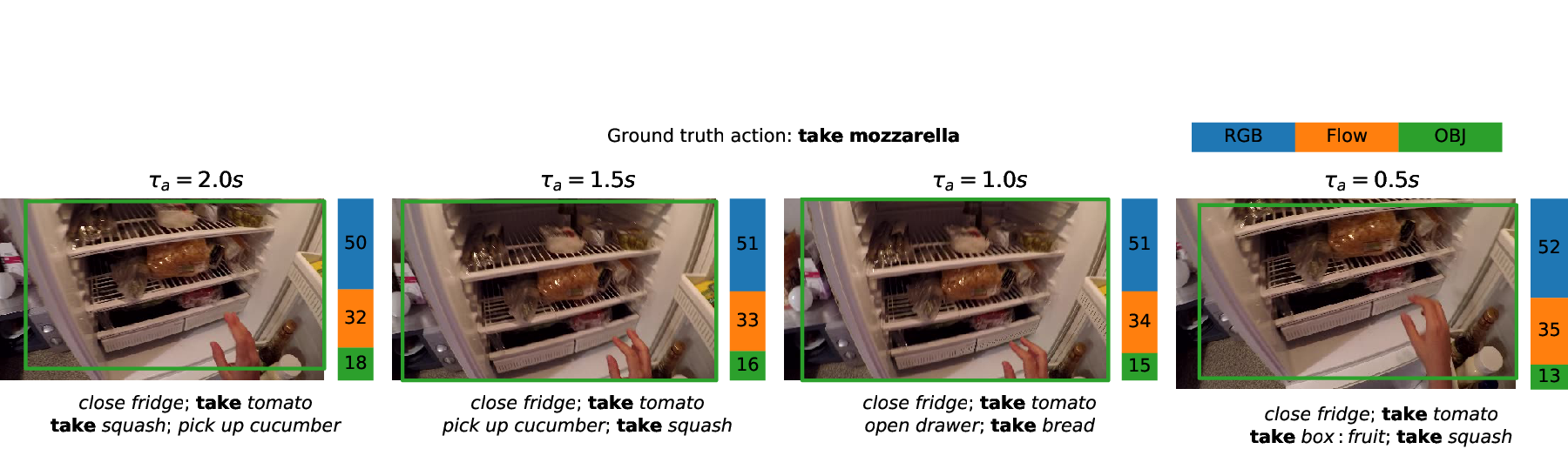}\\
	\includegraphics[width=\linewidth,clip=true,trim=0 0 0 190]{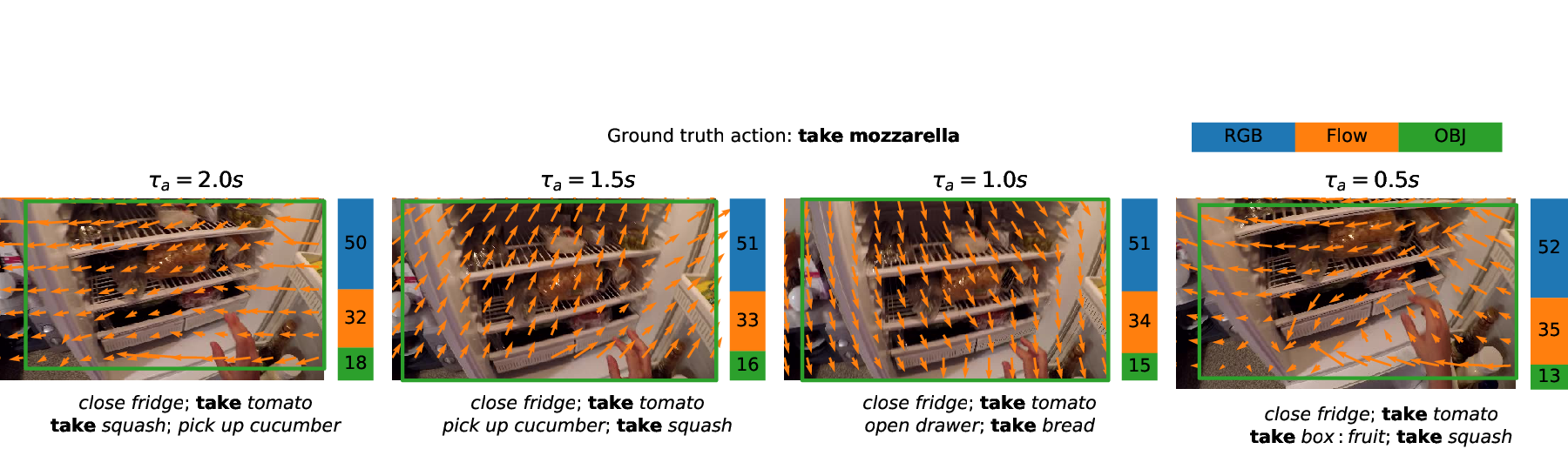}
	
	\includegraphics[width=\linewidth,clip=true,trim=0 64 0 60]{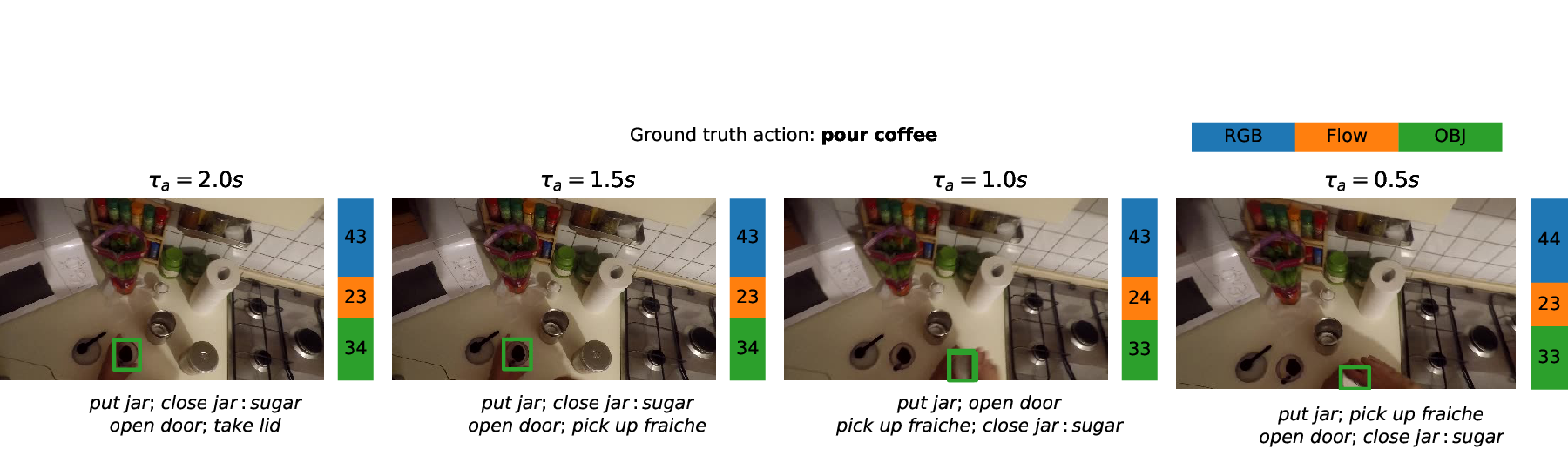}\\
	\includegraphics[width=\linewidth,clip=true,trim=0 0 0 190]{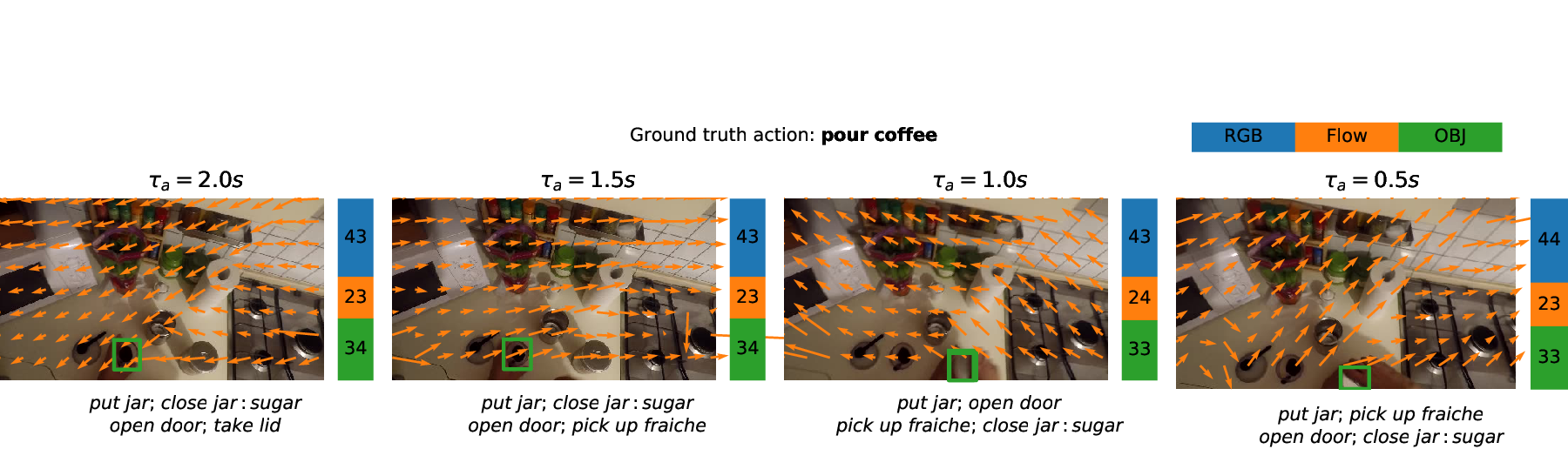}

	\caption{Failure action anticipation example qualitative results (best seen on screen).}
	\label{fig:qualitative_failure}
\end{figure*}

\section{Additional Qualitative Examples}
\label{sec:qualitative}
\figurename{~\ref{fig:qualitative_success}} reports qualitative results of three additional success action anticipation examples. For improved clarity, we report frames with and without optical flows for each example. In the top example, MATT assigns a small weight to the object branch as the contextual appearance features (i.e., RGB) are already enough to reliably anticipate the next actions. In the middle example object detection is fundamental to correctly anticipate ``put down spoon'', as soon as the object is detected. The bottom example shows a complex scene with many objects. The ability to correctly recognize objects is fundamental to anticipate certain actions (i.e., ``wash spoon''). The algorithm can anticipate ``wash'' well in advance. As soon as the spoon is detected ($\tau_a=2s$), ``wash spoon'' is correctly anticipated. Note that, even if the spoon is not correctly detected at time $\tau_a=0.5s$, ``wash spoon'' is still correctly anticipated.

\figurename~\ref{fig:qualitative_failure} reports three failure examples. In the top example, the model fails to predict ``adjust chair'', probably due to the inability of the object detector to identify the chair. Note that, when the object ``pan'' on the table is detected, ``take curry'' is wrongly anticipated. In the middle example, the algorithm successfully detects the fridge and tries to anticipate ``close fridge'' and some actions involving the ``take'' action, with wrong objects. This is probably due to the inability of the detector to detect ``mozzarella'', which is not yet appearing in the scene. In the bottom example, the method tries to anticipate actions involving ``jar'', as soon as ``jar'' is detected. This misleads the algorithm as the correct action is ``pour coffee''.

The reader is referred to the videos available at \textit{https://iplab.dmi.unict.it/rulstm/} for additional success and failure qualitative examples.

\bibliographystyle{IEEEtran}
\bibliography{egbib}

\end{document}